\documentclass[letterpaper]{article} 
\pdfoutput=1
\usepackage{aaai24}  
\usepackage{times}  
\usepackage{helvet}  
\usepackage{courier}  
\usepackage[hyphens]{url}  
\usepackage{graphicx} 
\usepackage{natbib}  
\usepackage{caption} 
\usepackage{algorithm}
\usepackage{algorithmic}

\usepackage{booktabs}
\usepackage{subfigure}
\usepackage{amsmath}
\usepackage{amssymb}
\usepackage{multirow}
\usepackage{color}
\usepackage{pifont}

\usepackage{newfloat}
\usepackage{listings}
\DeclareCaptionStyle{ruled}{labelfont=normalfont,labelsep=colon,strut=off} 
\floatstyle{ruled}
\newfloat{listing}{tb}{lst}{}
\floatname{listing}{Listing}
\usepackage{hyperref}
\usepackage{navigator}
\usepackage{nameref}
\nocopyright

\title{GroundVLP: Harnessing Zero-shot Visual Grounding from Vision-Language Pre-training and Open-Vocabulary Object Detection}
\author {
    Haozhan Shen\textsuperscript{\rm 1},
    Tiancheng Zhao\textsuperscript{\rm 2}\thanks{Corresponding author},
    Mingwei Zhu\textsuperscript{\rm 1},
    Jianwei Yin\textsuperscript{\rm 1}
}
\affiliations {
    \textsuperscript{\rm 1}Zhejiang University\\
    \textsuperscript{\rm 2}Binjiang Institute of Zhejiang University\\
    \{hz\_shen, zhumw, zjuyjw\}@zju.edu.cn,
     tianchez@zju-bj.com,
}

\usepackage{bibentry}

\begin{document}

\maketitle

\begin{abstract}

Visual grounding, a crucial vision-language task involving the understanding of the visual context based on the query expression, necessitates the model to capture the interactions between objects, as well as various spatial and attribute information. However, the annotation data of visual grounding task is limited due to its time-consuming and labor-intensive annotation process, resulting in the trained models being constrained from generalizing its capability to a broader domain. To address this challenge, we propose GroundVLP, a simple yet effective zero-shot method that harnesses visual grounding ability from the existing models trained from image-text pairs and pure object detection data, both of which are more conveniently obtainable and offer a broader domain compared to visual grounding annotation data. GroundVLP proposes a fusion mechanism that combines the heatmap from GradCAM and the object proposals of open-vocabulary detectors. We demonstrate that the proposed method significantly outperforms other zero-shot methods on RefCOCO/+/g datasets, surpassing prior zero-shot state-of-the-art by approximately 28\% on the test split of RefCOCO and RefCOCO+. Furthermore, GroundVLP performs comparably to or even better than some non-VLP-based supervised models on the Flickr30k entities dataset. Our code is available at \href{https://github.com/om-ai-lab/GroundVLP}{https://github.com/om-ai-lab/GroundVLP}.

\end{abstract}

\section{Introduction}

\begin{figure}[ht!]
    \centering
    \includegraphics[width=8cm]{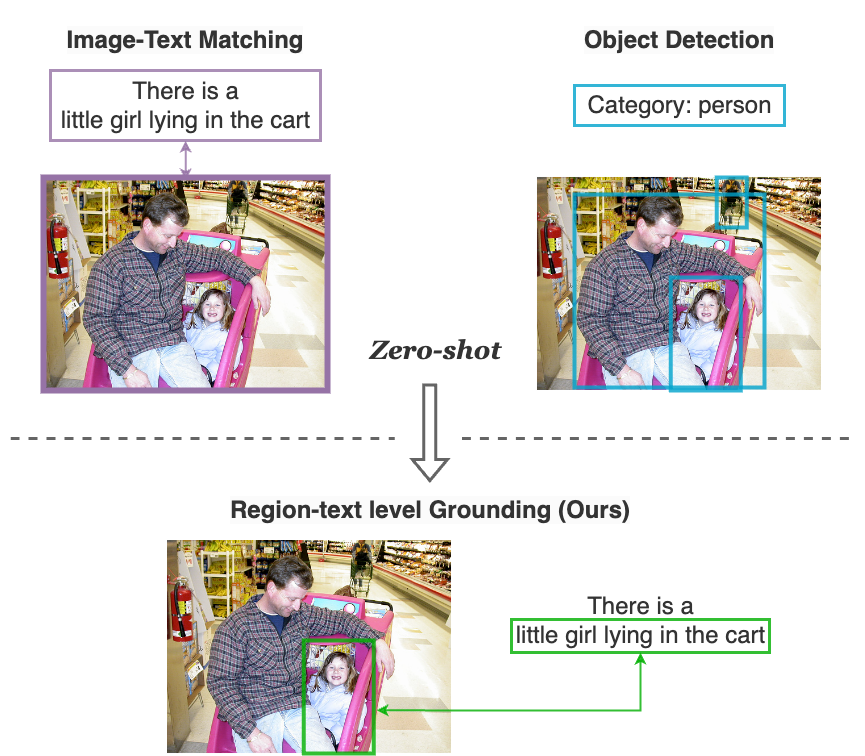}
    \caption{With the combination of existing models trained with image-text matching and object detection, we could conduct zero-shot visual grounding without fine-tuning on any additional supervised dataset. }
    \label{fig:introduce}
\end{figure}

Visual grounding seeks to pinpoint the image region described by a linguistic expression containing complex semantic information. It includes two typical tasks,  Referring Expression Comprehension (REC) and Phrase Grounding. REC aims to localize an object in an image given a textual referring expression, while phrase grounding seeks to ground every entity in the sentence to objects in the image. Generally, models are trained via task-specific datasets in the supervised setting~\cite{yu2018mattnet,liu2019knowledge,sun2021discriminative,yang2019fast,liao2020real,deng2021transvg} to perform visual grounding.

However, creating these task-specific datasets poses challenges due to their intricate annotation process. Annotating a query demands a detailed examination of object interactions and an understanding of various spatial and attribute information within the image. As a result, these datasets are limited in quantity, especially when compared to two other dataset types, as detailed in Table \ref{table:data compare}.  Undoubtedly, this finite amount of data restricts a model's adaptability to broader domain. Compared to the difficulty of obtaining visual grounding data, two alternative data types—image-text pairs and object detection\;(OD) data—are comparatively easier to obtain, as shown in Table \ref{table:data compare}. 

Recently, Vision-Language Pre-training\;(VLP) models, when trained on image-text pairs, have demonstrated impressive results in image-text matching\;(ITM)~\cite{zhang2021vinvl,kim2021vilt,li2021align}. Similarly, Open-Vocabulary object Detectors\;(OVD) trained with OD data have excelled in detecting specific categories~\cite{zareian2021open, gu2021open, zhou2022detecting}, as referenced in the top of Figure \ref{fig:introduce}. Therefore, a natural question arises: Can we harness the semantic comprehension of VLP and the category-specific detection prowess of OVD to perform visual grounding without any additional training, just as shown in the bottom of Figure \ref{fig:introduce}.

\begin{table}[h!]
  \begin{center}
  \resizebox{\hsize}{!}{
    \setlength\tabcolsep{3pt}
    \begin{tabular}{l c c c c } 
    \toprule
      \textbf{Data Type} &  &  \textbf{Name} & & \textbf{Total Size} \\
    \midrule
    \multirow{4}{*}{Visual Grounding}  &\vline & RefCOCO/+/g\cite{yu2016modeling, mao2016generation} &\vline & \multirow{4}{*}{$ \sim $ 220k} \\
     &\vline & CLEVR-Ref+~\cite{liu2019clevr} &\vline \\
        &\vline & Cops-Ref~\cite{chen2020cops} &\vline \\
        &\vline & Flicker30k~\cite{plummer2015flickr30k} &\vline \\
      \midrule
      
      \multirow{5}{*}{Image-Text Pair} &\vline & COCO~\cite{lin2014microsoft} &\vline & \multirow{5}{*}{$ \sim $ 6.2B}\\
      &\vline & Visual Genome~\cite{krishna2017visual} &\vline \\
      &\vline & SBU Caption~\cite{ordonez2011im2text} &\vline \\
        &\vline & CC3M/12M~\cite{changpinyo2021conceptual} &\vline \\
        &\vline & LAION400M/5B~\cite{schuhmann2021laion,schuhmann2022laion} &\vline \\
    \midrule
    
      \multirow{4}{*}{Object Detection} &\vline & COCO~\cite{lin2014microsoft} &\vline & \multirow{4}{*}{$ \sim $ 23.3M}\\
      &\vline & LVIS~\cite{gupta2019lvis} &\vline \\
      &\vline &  OpenImages~\cite{kuznetsova2020open}&\vline \\
        &\vline &  ImageNet-21K$^{\dagger}$~\cite{deng2009imagenet}&\vline \\
    \bottomrule
    \end{tabular}
    }
  \end{center}
  \caption{The list of some widely used datasets for three tasks. Size means the number of images included in the dataset.$^{\dagger}$ ImageNet, an image classification dataset, has been proven to be applicable to OD task by ~\cite{zhou2022detecting}. }
  \label{table:data compare}
\end{table}

In this paper, we introduce GroundVLP
, a novel method for zero-shot visual grounding tasks that encompasses both REC and phrase grounding. GroundVLP comprises three main components: (1) a VLP model that employs GradCAM to identify the image regions that are most semantically relevant to the given \textit{[expression]}\footnote{In this paper, \textit{[query]} refers to the query sentence provided by the grounding datasets and \textit{[expression]} refers to a specific object we need to ground. For REC, \textit{[expression]} is equivalent to \textit{[query]}. For phrase grounding, it denotes a certain entity phrase included in \textit{[query]}.}, (2) an OVD to detect the candidate objects, and (3) a fusion mechanism that combines the aforementioned two parts using a \textit{weighted grade} to select the answer judiciously. In contrast to its previous usage in the literature~\cite{li2021align, he2022vlmae}, we aggregate the GradCAM attention values only for visually recognizable words to optimize the modality mapping from text to image. Another significant difference is that we narrow down the candidate boxes to those belonging to a given object category with an OVD for to reduce noisy candidates compared to previous zero-shot methods~\cite{yao2021cpt, subramanian2022reclip}. The object category can be manually defined or predicted from the textual query using off-the-shelf NLP toolboxes such as Spacy~\cite{honnibal2015improved} or Stanza~\cite{qi2020stanza}.

We conduct main experiments on RefCOCO/+/g datasets for REC and Flickr30k Entities dataset for phrase grounding. GroundVLP outperforms all other zero-shot methods, which obtains an accuracy on average $\sim$18\% better than ReCLIP~\cite{subramanian2022reclip} on all splits of RefCOCO/+/g and $\sim$36\% better than CPT~\cite{yao2021cpt} on all splits of Flickr30k Entities. Experiment results further show that it performs on par or even better than some non-VLP-based supervised models on most of the test data. This outstanding performance indicates that we can tackle visual grounding tasks, which is traditionally constrained by limited annotations, using easily accessible data such as image-text pairs and pure object detection data. Additionally, we take ablation studies on each component of GroundVLP, demonstrating their effectiveness.

Our contributions could be summarized as:
(1) We propose a simple yet effective zero-shot method supporting both REC and phrase grounding, which achieves performance comparable to some non-VLP-based supervised models, demonstrating that visual grounding could be addressed using easily accessible data.
(2) We probe the cause of the decline in performance when not using the ground-truth category and discover inherent noise and bias on RefCOCO/+/g datasets.
(3) We conduct detailed ablation studies to verify the effectiveness of each proposed component and demonstrate the weak visual grounding capability of OVD.

\section{Preliminary}\label{preliminary}

\begin{figure}[htbp]
\centering
\subfigure[]{
\begin{minipage}[t]{0.4\linewidth}
\centering
\includegraphics[width=3.5cm]{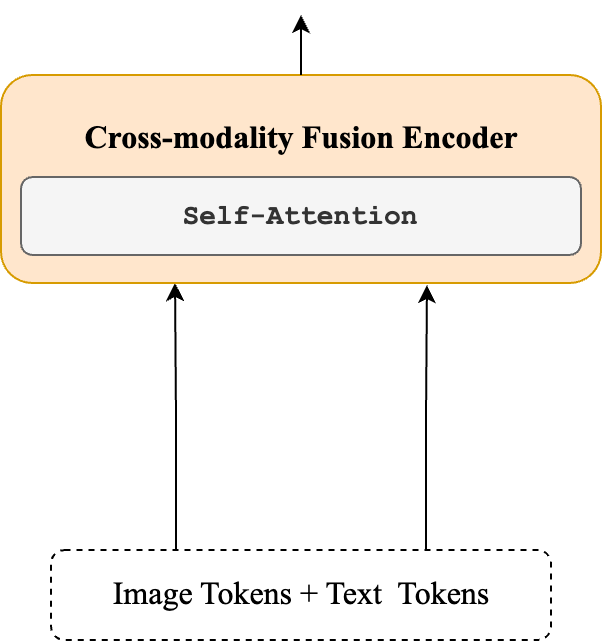}
\end{minipage}
}
\subfigure[]{
\begin{minipage}[t]{0.5\linewidth}
\centering
\includegraphics[width=3.5cm]{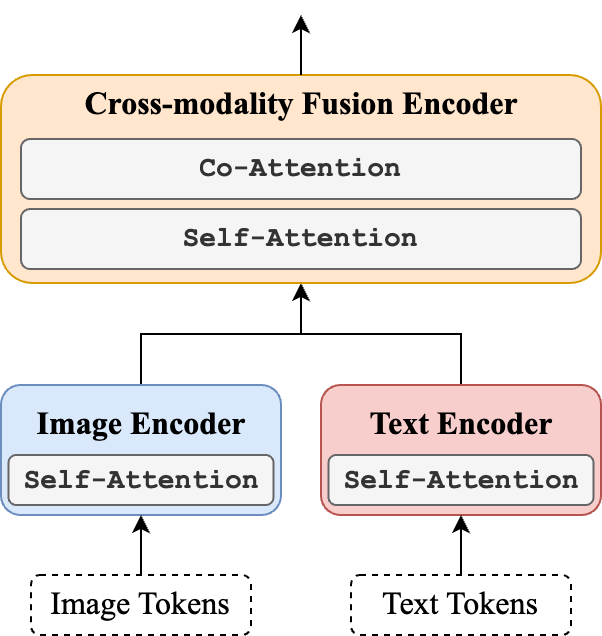}
\end{minipage}
}
\centering
\caption{Two types of attention architectures: \textbf{(a)} One-stream. \textbf{(b)} Two-stream. The two-stream model has an additional co-attention module in its cross-modality encoder compared to the one-stream model, which is used to interact with the information from two modalities.}
\label{fig:two architectures}
\end{figure}

There are two widely used attention modules: self-attention and co-attention, where the former employs the query(\textbf{Q}), key(\textbf{K}), and value(\textbf{V}) matrices created by the input sequence itself while the latter collects \textbf{K} and \textbf{V} from another sequence\cite{vaswani2017attention}. Existing VLP models can be roughly grouped into three types of architecture consisting of the aforementioned two attention modules: one-stream, two-stream, and dual-encoders. We mainly take the first two types into account and depict them in Figure. \ref{fig:two architectures}. Our usage of GradCAM~\cite{selvaraju2017grad} for these two attention architectures is described in detail here. 

Given a text-image pair, we input them into VLP and define \textit{T}, \textit{I} as the number of text and image input tokens respectively. The attention map of a certain layer, denoted as \textbf{A}, can be computed by the product of \textbf{Q} and \textbf{K} with furthermore post-processes, as defined specifically in the equation: $\textbf{A} = \textit{softmax}(\frac{\textbf{Q} \cdot \textbf{K}^\top}{\sqrt{d_{h}}})$, where \textbf{Q} $\in$ $ \mathbb{R}^{N_h \times s \times d_h }$, \textbf{K} $\in$ $ \mathbb{R}^{N_h  \times q \times d_h }$ and  \textbf{A} $\in$ $\mathbb{R}^{N_h \times s \times q }$. $N_h$ is he number of attention heads of multi-head attention, \textit{$d_h$} means the dimension of hidden states, and \textit{s}, \textit{q} are assigned different values in different architecture:

\newlength{\widest}
\settowidth{\widest}{}
\begin{equation}
      (\textit{s}, \textit{q}) =
     \begin{cases}
       \makebox[\widest][l]{(\textit{T}+\textit{I}, \textit{T}+\textit{I}) } & \quad \text{one-stream}\\
       (\textit{T}, \textit{I}) \qquad  & \quad \text{two-stream}
      \end{cases}
\end{equation}

where for the two-stream architecture, we compute \textbf{A} in the co-attention module of the fusion encoder in which \textbf{Q} is from the language encoder and \textbf{K} from the image. Next, we compute the gradients map via back propagation:$\nabla{\textbf{A}} = (\frac{\partial{L_{itm}}}{\partial{\textbf{A}}})^+$, where $L_{itm}$ represents the VLP model's output value of the ITM head, and we remove the negative contributions. Finally, the result map ${\textbf{G}}$ $\in$ $\mathbb{R^{\textit{s} \times q}}$ is given by:
$${\textbf{G}} = \mathbb{E}_{h} (\nabla{\textbf{A}} \odot \textbf{A})$$where $\mathbb{E}_{h}$ is the average calculation across heads dimension and $\odot$ means  element-wise multiplication.

To sum up, for a given layer in the encoder, we can obtain a map  \textbf{G} $\in$ $\mathbb{R^{\textit{s}  \times \textit{q}}}$ via GradCAM.

\section{The Proposed Method}

\begin{figure*}[h]
    \centering
    \includegraphics[width=16cm]{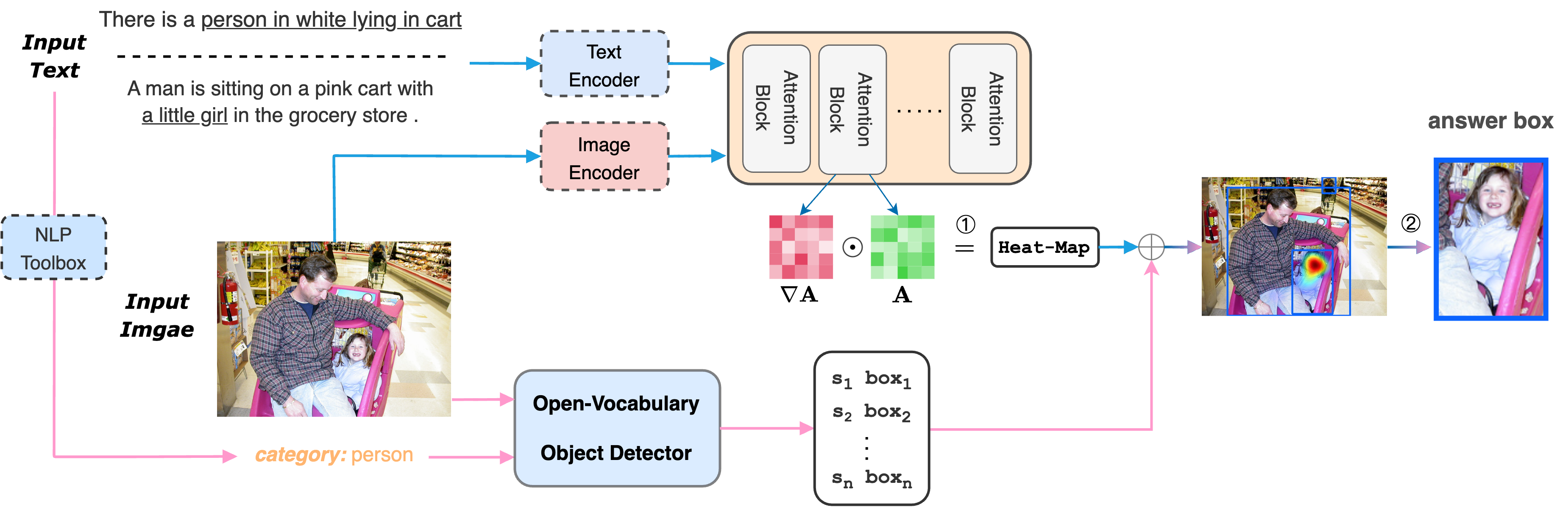}
     \caption{Overview of GroundVLP. The \underline{underline} words indicate the \textit{[expression]}. The \textcolor[RGB]{27,161,226}{blue} arrow lines denote the procedures involving GradCAM, and the \textcolor[RGB]{255,153,204}{pink} lines are related to the open-vocabulary object detector. The dashed line indicates that the module may not exist. $\left\{(s_k,\textbf{box}_k)\right\}_1^n$ represents an array of confidence scores and bounding boxes detected by OVD based on the given category. The symbol $\odot$ means element-wise multiplication of attention map and gradients,and  $\oplus$ means fusing the instances detected by OVD and heat-map. At location \ding{172}, we utilize \textit{visual-word attention aggregation}, while at location \ding{173}, we employ \textit{weighted grade} to select the answer box.}
    \label{fig:overview}
\end{figure*}

Figure. \ref{fig:overview} demonstrates an overview of GroundVLP. We first obtain \textbf{G} via GradCAM. Then we crop its size and apply our proposed \textit{visual-word attention aggression} to obtain the heat map of VLP. For the OVD module, we input the category of the referring target into it to receive \textit{n} instances including boxes and confidence scores. Finally, we fuse two parts to calculate the \textit{weighted grades} of each instance and output the one with the highest grade.

\subsection{Generating a Heat-map for VLP}\label{Generating a heat-map for VLP}
\begin{figure*}[htbp]
\centering
\subfigure[Use $\mathtt{[CLS]}$ merely]{
\begin{minipage}[t]{0.31\linewidth}
\centering
\includegraphics[width=4.5cm]{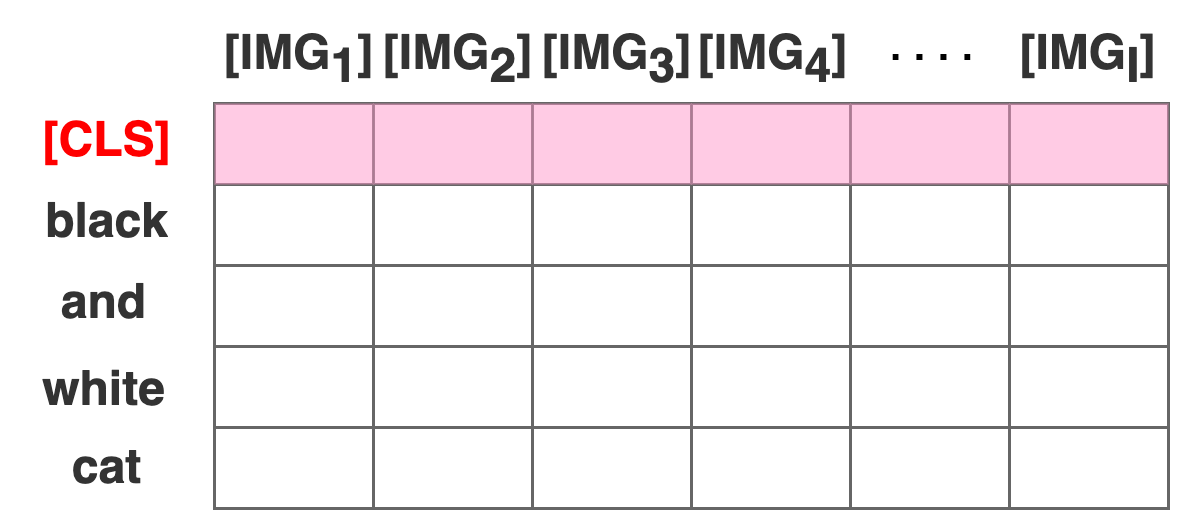}
\end{minipage}
}
\subfigure[Use all text tokens]{
\begin{minipage}[t]{0.31\linewidth}
\centering
\includegraphics[width=4.5cm]{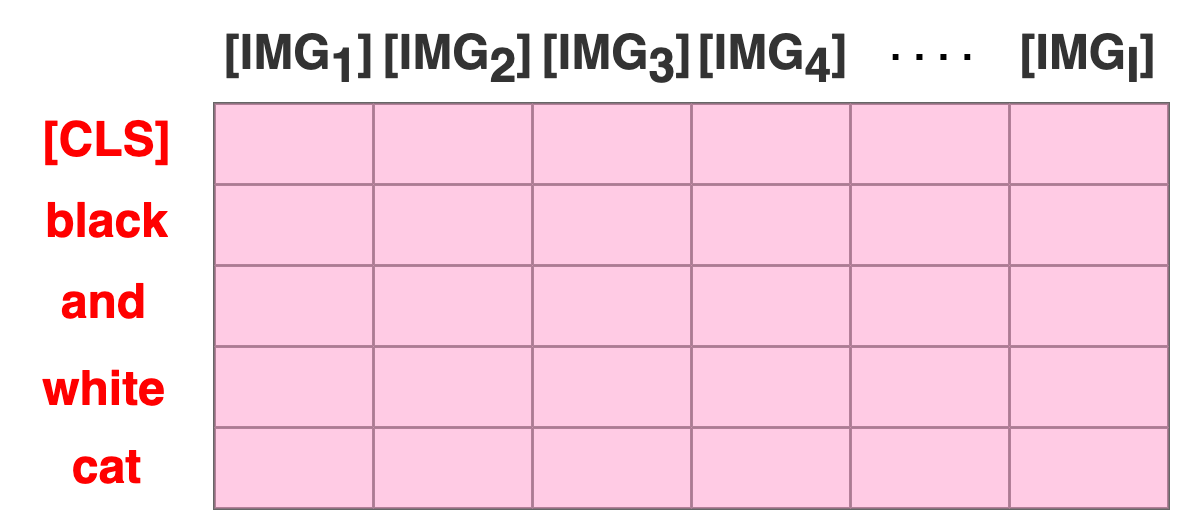}
\end{minipage}
}
\subfigure[\textbf{Ours}]{
\begin{minipage}[t]{0.31\linewidth}
\centering
\includegraphics[width=4.5cm]{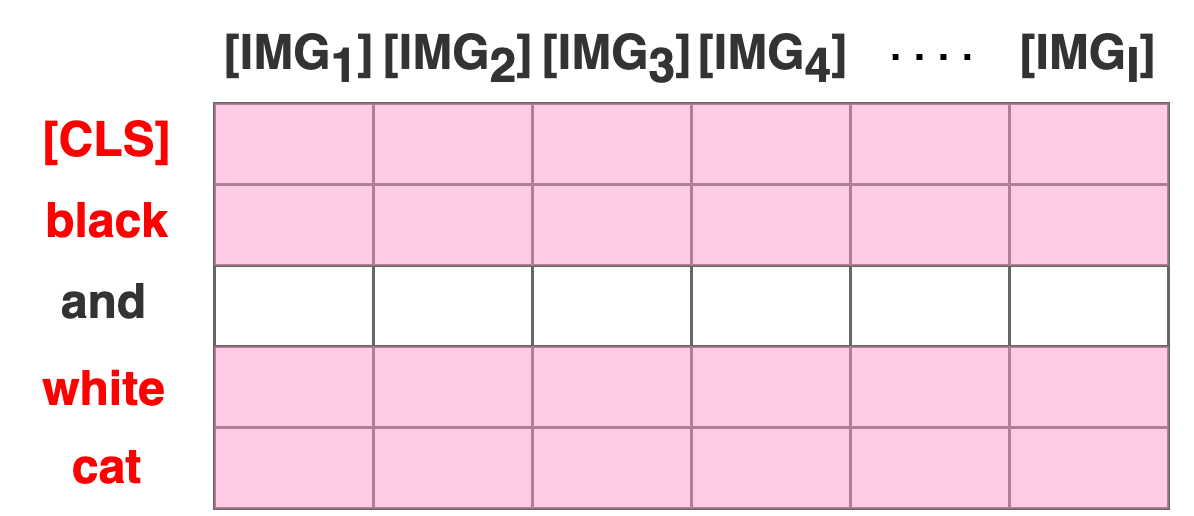}
\end{minipage}
}
\centering
\caption{The difference between the proposed visual word attention aggregation and prior methods}
\label{fig:three methods}
\end{figure*}

To ground the \textit{[expression]}, we prompt the \textit{[query]} and feed it into the model along with the image to obtain \textbf{G} by GradCAM. Then we can generate a heat map for VLP. First, we crop \textbf{G} $\in \mathbb{R^{\textit{s} \times \textit{q}}}$ to $\textbf{G}^{\prime} \in \mathbb{R}^{\textit{T} \times \textit{I}}$, where $\textbf{G}^{\prime}$ denotes the influence of the image tokens on each text token:
\begin{equation}
      \textbf{G}^{\prime} =
      \begin{cases}
      \makebox[\widest][l]{$\textbf{G}[i, j]_{1\le i\le T}^{I\le j\le T+I}$} & \quad \text{one-stream}\\
                            \textbf{G} \qquad \qquad \quad & \quad \text{two-stream}
      \end{cases}
\end{equation}                   
Next, $\textbf{G}^{\prime}$ needs to be further squeezed to $\tilde{\textbf{G}}$ $\in$ $\mathbb{R}^I $ so that it can represent the connections between the whole \textit{[expression]} and each image token.  To this end, we propose \textit{visual-word attention aggregation}, distinguished from previous methods which are implemented by either using the row corresponding to the $\mathtt{[CLS]}$ token~\cite{subramanian2022reclip} directly or averaging the scores of rows across all text tokens~\cite{li2021align,he2022vlmae}, as shown in Figure. \ref{fig:three methods}.

\textbf{Visual-Word Attention Aggregation:} An \textit{[expression]} consists of words with various part-of-speech (POS) tags, where some words can be easily mapped to a specific image region. For instance, for the phrase ``\textit{black and white cat}", it is easy to locate ``\textit{black}", ``\textit{white}" and ``\textit{cat}", but ``\textit{and}" is less clear. Thus, we conjecture that VLP models will also perform well when mapping visually recognizable words to the image.

We define $\mathcal{V}$ as a set of  POS tags, including nouns, adjectives, verbs, proper nouns, and numerals, which are relatively easy to visualize. An off-the-shelf NLP processing toolbox is utilized to parse the POS tag of each word in the \textit{[expression]} and only those whose tag is included in $\mathcal{V}$ will remain. We denote $\mathcal{W}$ as the set of \textit{[expression]}'s text tokens. Filtered by $\mathcal{V}$, original set $\mathcal{W}$ is cut down to $\mathcal{W}^\prime $, and we further add $\mathtt{[CLS]}$ token into $\mathcal{W}^\prime $ when conducting REC because of its general representation of all the tokens. After that $\tilde{\textbf{G}}$ $\in$ $\mathbb{R}^I $ can be calculated as:
$$\tilde{\textbf{G}} = \mathbb{E}_{t}(\textbf{G}^\prime), t\in \mathcal{W}^\prime$$
where $\mathbb{E}_{t}$ means the average calculation across text dimension and we only calculate the text tokens included in $\mathcal{W}^\prime$.

We then reshape $\tilde{\textbf{G}}$ to \textbf{H} $\in$ $\mathbb{R}^{\textit{h} \times \textit{w}}$ having the same size as the input image. VLP can be divided into region-based and end-to-end by whether relying on an external object detector to obtain the visual inputs, which should be applied by different reshaping patterns. We introduce both types into GroundVLP and use different patterns for each type.

\textbf{Heat-map for Region-Based Models:} Region-based VLP models transform the image into a set of region proposals as visual features. Thus, we select a subset of image tokens with high attention values and superimpose their values onto the corresponding image regions to generate the heat-map.

To do this, we sort each element of $\tilde{\textbf{G}}$ in descending order according to their attention values and select the top \textit{m} tokens among them, defined as $\left\{ (v_k, \textbf{b}_k) \right\}_1^m$, where $v_k$ and $\textbf{b}_k = (x_{k1},y_{k1},x_{k2},y_{k2})$ are the corresponding attention value and coordinates of the proposal. The heat map of  the region-based model $\textbf{H}_{R}$ is computed as:
\begin{equation}
      \textbf{H}_k[i, j] = 
      \begin{cases}
      \makebox[\widest][l]{$v_k$}  & \quad (i, j) \in \textbf{b}_k  \\
                            0    & \quad \text{otherwise}
      \end{cases}
\end{equation}

$\textbf{H}_{R} = \sum_{k=1}^{m}\textbf{H}_k$, where $\textbf{H}_k$ is a $h\times w$ matrix, $R$ represents Region-based, and the sum calculation for $\textbf{H}_k$ is implemented by element-wise addition.

\textbf{Heat-map for End-to-End Models:} End-to-end VLP models process visual input as a set of patch embeddings with vision-transformer~\cite{dosovitskiy2020image}. Its image tokens are a series of image patches. Following ~\cite{li2021align,he2022vlmae}, we employ a bicubic interpolation on $\tilde{\textbf{G}}$ to reshape it to $\textbf{H}_{E}$ $\in$ $\mathbb{R}^{\textit{h} \times \textit{w}}$, where $E$ represents End-to-end.

\subsection{Fusion with Open-Vocabulary Detectors}
Having obtained the heat-map \textbf{H} $\in$ $\mathbb{R^{\textit{h}  \times \textit{w}}}$, we proceed to generate a set of candidate boxes, calculate the \textit{weighted grades} of regions enclosed by each one, and output the box with the highest grade. We focus on the boxes belonging to the predetermined category merely, simplifying the selection of the final answer box as it reduces the number of candidate boxes. In view of a user should have a specific category in mind when it comes to real-world applications, we employ the ground-truth category during the REC task to mimic the user's input. Meanwhile, we also present an alternative manner to extract the target unit from the \textit{[expression]} as the predicted category when no category is provided.

\begin{figure}[h]
    \centering
    \includegraphics[width=6cm]{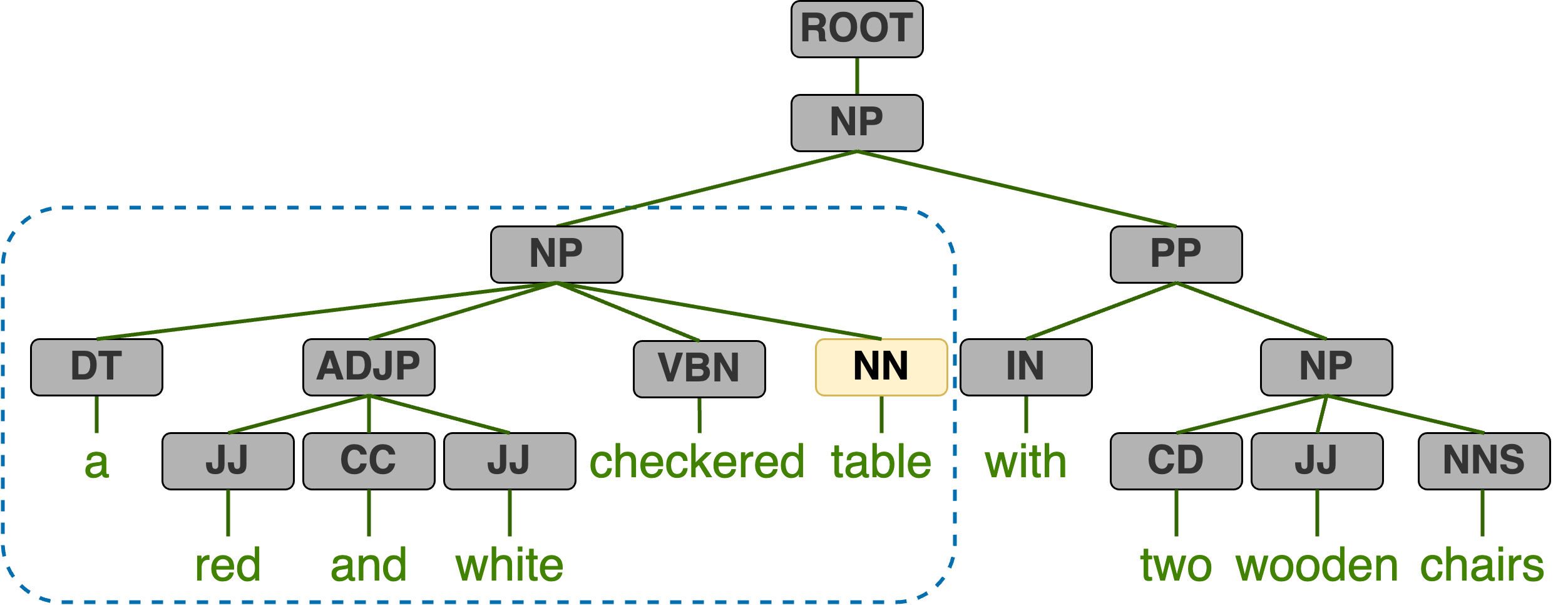}
    \caption{An example of category extract. The original query is ``\textit{a  red and white checkered table with two wooden chairs}". The \textcolor[RGB]{0,110,175}{blue} dash rectangle indicates the bottom-left NP node and \textcolor[RGB]{255,215,0}{yellow} NN node is the target unit.}
    \label{fig:category extract}
\end{figure}

\textbf{Category Extraction:} Inspired by ~\cite{sun2021discriminative}, we exploit an NLP toolbox to extract the target unit of the \textit{[expression]} as the predicted category. Specifically, a dependency tree of the \textit{[expression]} is generated by the NLP toolbox, and its rightmost Normal Noun\,(NN) node of the far left-bottom Noun Phrase\,(NP) node is viewed as the predicted category.  An example is illustrated in Figure. \ref{fig:category extract}.

Furthermore, we map the extracted predicted categories to the class vocabulary of the evaluation dataset to conform to the ground-truth categories. Let $\mathcal{C}$ be the set of classes vocabulary of the evaluation dataset, \textit{$c_i$} $\in$ $\mathcal{C}$ be one of the classes, and \textit{$c^p$} be the extracted predicted category. Next, we employ CLIP~\cite{radford2021learning} to embed \textit{$c_i$} and $c^p$ as $\textbf{e}_{c_i}$ $\in$ $\mathbb{R^{\textit{D}}}$ and $\textbf{e}^p$ $\in$ $\mathbb{R^{\textit{D}}}$ for projecting them into an uniform embedding space, where \textit{D} denotes the dimension of  CLIP embedding. As per CLIP, the prompt ``\textit{a photo of}" is added as a prefix to $c_i$ and $c^p$ before embedding them. We then define $c^{map}$ as the category mapped from $c^p$ to $\mathcal{C}$ and $sim_i$ as the similarity between $c^p$ and $c_i$, which are given as follow:

\begin{equation}
\label{eq:map category}
    sim_i=
P(c^{map}=c_i\mid c^p ) = \frac{exp(\textbf{e}_{c_i}^\top\textbf{e}^p)}{\sum_{j=1}^{\lvert \mathcal{C} \rvert}exp(\textbf{e}_{c_j}^\top\textbf{e}^p)}
\end{equation}
$$c^{map} = c_i,\  i = argmax_i(sim_i)$$

Through this equation, each $c^p$ can be mapped to a specific class. For instance, ``\textit{table}" in Figure. \ref{fig:category extract} will be mapped to ``\textit{dining table}" when using COCO vocabulary.

\textbf{Generating Candidate Boxes:} Given an image $\mathcal{I}$, the predetermined category $c$ and a score threshold $\theta$, an OVD is allowed to detect \textit{n }instances:
\begin{align}
    \left\{(s_k,\textbf{box}_k)\right\}_1^n &=\ 
    \textbf{OVD}\;(c,\ \theta,\ \mathcal{I}\,) \\
    \textbf{box}_k\quad &=\  (x_{k1},y_{k1},x_{k2},y_{k2}) \\
    s_k\quad\ \ \   &=\ P(o_k\in c\mid \mathcal{I}\,),\ s_k>\theta
\end{align}
where $o_k$ represents the object entity included in $\textbf{box}_k$, and $s_k$ denotes the confidence score that $o_k$ belongs to $c$ , which should be greater than $\theta$.  

\textbf{Weighted Grade:} A crucial challenge we face now is ascertaining the value of $\theta$. If it is set too low, the superfluous boxes, which do not belong to $c$, may be included. Conversely, if $\theta$ is too high, certain boxes belonging to $c$ may be missed. For this problem, we raise a formula that set a relatively low threshold and let $s_k$ be a weight of $r_k$,  where  $r_k$ is defined as the total heat-map value of the region enclosed by $\textbf{box}_k$. Through it, we balance both not omitting boxes that pertain to $c$ and preventing low-scoring boxes from disturbing the result.  Finally, we calculate $g_k$ which represents a \textit{weighted grade} of $(s_k,\textbf{box}_k)$ and output $\textbf{box}_{pred}$ with the highest grade as the prediction of GroundVLP:

\begin{align}
    r_k &= \sum_{i=x_{k1}}^{x_{k2}}\sum_{j=y_{k1}}^{y_{k2}}\textbf{H}[i,\ j]  \\ 
    g_k &= \frac{1}{A_k^\alpha}\cdot s_k\cdot r_k \\
    \textbf{box}_{pred} &= \textbf{box}_k,\  k = argmax_k(g_k)
\end{align}

where $A_k$ means the area of $\textbf{box}_k$ used to avoid the tendency to choose boxes with large areas and $\alpha$ is a hyperparameter.

\section{Experiments}

\subsection{Datasets}
\label{dataset}
\textbf{Referring Expression Comprehension:}\quad We adopt three widely used datasets: RefCOCO, RefCOCO+~\cite{yu2016modeling} and RefCOCOg~\cite{mao2016generation}. RefCOCO and RefCOCO+ are both split into validation, testA, and testB sets, where testA generally contains queries with persons as referring targets and testB contains other types. RefCOCO is described by  more spatial information compared to RefCOCO+, whereas RefCOCO+ contains queries using more appearance-related words instead. In contrast, RefCOCOg has longer and more detailed expressions than the other two datasets. 

\noindent \textbf{Phrase Grounding:} We adopt Flickr30k entities dataset~\cite{plummer2015flickr30k} for the task and evaluate the performance in terms of Recall@1, 5. On Flickr30k, a sentence contains several phrases that need to be grounded, each of which may correspond to multiple bounding boxes. Hence, previous researches propose two protocols named ANY-BOX and MERGED-BOX by MDETR~\cite{kamath2021mdetr}. In our evaluation, we use the ANY-BOX protocol.

\begin{table*}[h!]
\footnotesize
  \begin{center}
    \begin{tabular}{l c c c c c c c c c c} 
    \toprule
       & \multicolumn{3}{c}{RefCOCO} & & \multicolumn{3}{c}{RefCOCO+} & & \multicolumn{2}{c}{RefCOCOg}\\
      \textbf{Model} & \textbf{val} & \textbf{testA} & \textbf{testB} & & \textbf{val} & \textbf{testA} & 
      \textbf{testB} & & \textbf{val} & \textbf{test}\\
    \midrule
      Supervised SOTA~\cite{yan2023universal} & 92.64 & 94.33 & 91.46& \vline & 85.24 & 89.63 & 79.79 & \vline & 88.73 & 89.37\\
      \midrule
      \textbf{Supervised method w/o VLP} & &  & & \vline &  &  &  & \vline &  & \\
      NMTree~\cite{liu2019learning} &76.41 & 81.21 & 70.09& \vline & 66.46 & 72.02 & 57.52 & \vline & 65.87 & 66.44 \\
      TransVG~\cite{deng2021transvg} & 81.02 & 82.72 &78.35 & \vline & 64.82 & 70.70 & 56.94 & \vline & 68.67  & 67.73 \\
    \midrule
     \textbf{Other zero-shot} & &  & & \vline &  &  &  & \vline &  & \\
      CPT-Seg w/ VinVL~\cite{yao2021cpt} & 32.2 & 36.1 & 30.3 & \vline & 31.9 & 35.2 & 28.8 & \vline & 36.7 & 36.5\\
      ReCLIP w/ relations~\cite{subramanian2022reclip} & 45.78 & 46.10 & 47.07 & \vline & 47.87 & 50.10 & 45.10 & \vline & 59.33 & 59.01 \\
    \midrule
      \textbf{Ours using predicted category} & & & & \vline & & & & \vline \\
      GroundVLP w/ ALBEF & 52.58 & 61.30 & 43.53 & \vline & 56.38 & 64.77 & 47.43 & \vline & 64.30 & 63.54\\ 
      GroundVLP w/ VinVL & \underline{59.05} & \underline{69.21} & 48.71 & \vline & 61.80 & 70.56 & 50.97 & \vline & 69.08 & 68.98\\
    \midrule
      \textbf{Ours using ground-truth category}& & & & \vline & & & & \vline \\
      GroundVLP w/ ALBEF & 58.22 & 65.48 & \underline{49.75}& \vline &\underline{63.57} & \underline{72.02} & \underline{53.57} & \vline & \underline{69.63} & \underline{69.15}\\ 
      GroundVLP w/ VinVL & \textbf{65.01} & \textbf{73.50} & \textbf{55.01}& \vline & \textbf{68.87} & \textbf{78.10} & \textbf{57.31} & \vline & \textbf{74.73} & \textbf{75.03}\\
    \bottomrule
    \end{tabular}
  \end{center}
  \caption{Accuracy (\%) on referral expression comprehension datasets. We show both the results of GroundVLP using the predicted category and that using the ground-truth category. The best zero-shot accuracy in each column is in \textbf{bold}, and the second best is \underline{underlined}. Supervised SOTA refers to UNINEXT \cite{yan2023universal}.}
  \label{table:main results rec}
\end{table*}

\subsection{Implementation Details}\label{implement}

\textbf{Selected VLP models and Prompt Templates:} $\mathcal{I}, \mathcal{T}$ is defined as the input image and text. We introduce a typical model from both region-based and end-to-end: VinVL~\cite{zhang2021vinvl} and ALBEF~\cite{li2021align}. We adopt VinVL-Large and ALBEF-14M as the checkpoints for two models. The input format of VinVL is a triple tuple $\left\{ \textbf{\textit{w}},\ \textbf{\textit{q}}\ ,\ \textbf{\textit{v}} \right\}$ and can be formed as two types.
\footnote{We refer readers to \nameref{appendix:vinvl input} to learn the details of the two formats.}.
We adopt the VQA-resemble and prompt the query to adapt it to this format. Specifically, we let $\mathcal{T}$ = ``\textit{there\ is a\ [query]?}" on REC task and $\mathcal{T}$ = ``\textit{[query]?}" on phrase grounding, and let \textbf{\textit{q}} always be ``\textit{yes}". For ALBEF, we prompt  $\mathcal{T}$ = `\textit{`there\ is\ a\ [query].}" on REC task and $\mathcal{T}$ = ``\textit{[query].}" on phrase grounding, where \textit{[query]} denotes the query sentence provided by the datasets. 

\textbf{GradCAM Layer:} For ALBEF, we use the $3^{rd}$  layer of the cross-modality encoder for GradCAM. For VinVL, we use the $20^{th}$ layer of the cross-modality encoder and select \textit{m} = 7. All setting is based on tuning on the RefCOCOg validation dataset.

\textbf{Methods for Category Prediction:} We employ Stanza~\cite{qi2020stanza} to extract predicted category. When testing on RefCOCO/+/g, we map the predicted category to the COCO class via equation \ref{eq:map category}. For Flickr30k entities, given that its ground-truth category is slightly abstract\footnote{They are: \textit{people}, \textit{clothing}, \textit{bodyparts}, \textit{animals}, \textit{vehicles},\textit{ instruments}, \textit{scene}, and \textit{other}.}, we use the predicted category directly. Besides, in order to detect a person entity better, we set $c$ = $\left\{c^p,\ person\right\}$ if the cosine similarity of $\textbf{e}^p$ and $\textbf{e}^{p*}$ is greater than 0.9, where $\textbf{e}^{p*}$ denotes the textual CLIP embedding of  ``\textit{a photo of person}".

\textbf{Selected Open-vocabulary Detector:} We choose Detic~\cite{zhou2022detecting} as our open-vocabulary detector (OVD). Other OVDs can also be considered~\cite{li2022grounded,zhao2022omdet}. For REC, we set $\alpha$ = 0.5, $\theta$ = 0.15 when using ground-truth category and $\theta$ = 0.3 for predicted category. For phrase grounding, we set $\alpha$ = 0.25 and $\theta$ = 0.15. If Detic detects no box, we use all proposals as candidate boxes instead. For RefCOCO/+/g, we adopt the proposals from MAttNet~\cite{yu2018mattnet}.  For Flickr30k entities, we use all proposals detected by Detic.

\textbf{Compared Baseline:} We choose two previous zero-shot REC methods to compare-- ReCLIP~\cite{subramanian2022reclip} and CPT~\cite{yao2021cpt}. CPT masks the regions of each proposal with different colors and predict the color word in " \textit{[query] is in} $\mathtt{[MASK]}$ \textit{color}", while ReCLIP scores each proposal by using the contrastive scoring ability of CLIP. Furthermore, we construct a CPT-adapted baseline for phrase grounding to compare. A \textit{[query]} on Flickr30k contains $N$ phrases, denoted as $\left\{[expression]_i\right\}_1^N$. We then copy $[query]$ with $N$ times and add ``\textit{where\ [expression]$_i$ is\  in [MASK] color}"  after $i^{th}$ duplication. An example is shown in \nameref{appendix:cpt-adapted}. We use all proposals detected by Detic and colored blocks~\cite{yao2021cpt} for CPT-adapted.

\begin{table}[h!]
\footnotesize
  \begin{center}
  \resizebox{\hsize}{!}{
  \setlength\tabcolsep{3pt}
    \begin{tabular}{l c c c c c } 
    \toprule
       & \multicolumn{2}{c}{Flickr30k val} & & \multicolumn{2}{c}{Flickr30k test} \\
      \textbf{Model} & R@1 & R@5 & & R@1 & R@5 \\
    \midrule
     \textbf{Supervised method  with VLP} & & &  \vline & &   \\
      MDETR-ENB5~\cite{kamath2021mdetr} & 83.6& 93.4 & \vline & 84.3 & 93.9  \\
      GLIP-L~\cite{li2022grounded} & 86.7& 96.4 &  \vline & 87.1 & 96.9 \\
    \midrule
     \textbf{Supervised method w/o VLP} & & & \vline & &  \\
      ZSGNet~\cite{sadhu2019zero} & -& - &  \vline & 63.39 & - \\
      BAN~\cite{kim2018bilinear} & -& - & \vline & 69.69 & 84.22 \\
    \midrule
      \textbf{Zero-shot method} & & & \vline & & \\
      CPT-adapted~\cite{yao2021cpt} & 27.06& 61.78 &  \vline & 27.57 & 61.55  \\
      GroundVLP w/ ALBEF(\textbf{Ours}) & 63.76 & \textbf{75.02} &  \vline & 63.89 & \textbf{74.80} \\ 
      GroundVLP w/ VinVL(\textbf{Ours}) & \textbf{63.89} & 74.53 & \vline & \textbf{64.19} & 74.57 \\
    \bottomrule
    \end{tabular}
    }
  \end{center}
  \caption{Accuracy (\%) on the Flickr30k entities dataset. We compare GroundVLP with VLP-based, non-VLP-based supervised methods and prior zero-shot method. The best zero-shot accuracy in each column is in \textbf{bold}. Note that we only use the predicted category during this task.}
  \label{table:main results pg}
\end{table}

\subsection{Main Results}

\textbf{Referring Expression Comprehension\footnote{Case study is shown in \nameref{case study}.}:} Table \ref{table:main results rec} shows the results on RefCOCO/+/g. GroundVLP outperforms other zero-shot methods, especially in the testA split of RefCOCO and RefCOCO+. When using the ground-truth category, GroundVLP is comparable or superior to some non-VLP-based supervised models. 

However, it is also noted that there is an inevitable decline in performance when using the predicted category compared to the case of using the ground truth. It can be attributed to several factors: \textbf{(1)} Unclear referring targets: the word of the target unit in the query may not clearly indicate the referring target. For instance, the query ``\textit{black hat}" indicates a person wearing a black hat, and the word ``\textit{hat}" will be extracted. However, ``\textit{hat}" cannot be mapped to ``\textit{person}" exactly, leading to mistakes. \textbf{(2)} Undisciplined grammar: there is a part of coarse queries on RefCOCO/+/g, where the NLP toolbox cannot extract the target unit accurately. For example, ``\textit{woman red coat}", an undisciplined expression of ``\textit{woman in red coat}" or  ``\textit{woman wearing red coat}", will cause the NLP toolbox to regard ``\textit{woman}" as a noun adjective used to describe ``\textit{red}" and treat ``\textit{coat}" as the target instead. \textbf{(3)} No target in query: there are a few queries consisting of pure spatial information, not containing the referring target (e.g. ``\textit{left}", ``\textit{the closest to you}"). We argue that these datasets inherently include bias and noise, which makes it difficult to accurately map the predicted category to a COCO class, resulting in a decline in performance. We take a more detailed demonstration in \nameref{noise}
.

\textbf{Phrase Grounding:} Table \ref{table:main results pg} shows the results on the Flickr30k entities dataset. GroundVLP is far ahead of CPT for the R@1 score, outperforming it by 38.29\% and 37.79\% in the val and test split. Moreover, GroundVLP performs comparably to or even better than some non-VLP-based supervised approaches. Finally, it is noticed that we only use predicted category for phrase grounding, demonstrating the effectiveness of the proposed fusion method for grounding tasks and the disciplined expression on the Flickr30k dataset that is beneficial for our predicted category extraction.

\begin{table}[h!]
  \begin{center}
  \resizebox{\hsize}{!}{
    \setlength\tabcolsep{4pt}
    \begin{tabular}{l c c c c} 
    \toprule
      \textbf{Model} & RefCOCO & RefCOCO+ & RefCOCOg & \textit{Difference}\\
    \midrule
     \textbf{GroundVLP} & & & &\\
      \quad w/ TCL & 59.59& 64.96 & 70.92 & \textit{5.37}\\
      \quad w/ PTP& 56.81& 61.17 & 68.53 & \textit{4.36}\\
      \quad w/ Lxmert& 63.14& 60.99 & 67.99& \textit{-2.15}\\
    \bottomrule
    \end{tabular}
    }
  \end{center}
  \caption{Accuracy (\%) of using other VLP models. We report the difference of the score on RefCOCO+ minus that on RefCOCO for each model in column \textit{Difference}. All datasets in the table indicate their val split and all results on RefCOCO/+/g are obtained by using the ground-truth category.}
  \label{table:results other models}
\end{table}

\subsection{Extending to Other VLP Models}

In order to verify the versatility of our method, we further incorporate more VLP models into GroundVLP. We present the results using TCL~\cite{yang2022vision}, PTP\~cite{wang2022position} and Lxmert~\cite{tan2019lxmert} in Table \ref{table:results other models}, among which TCL and PTP belong to end-to-end and Lxmert belongs to region-based.The brief description and implementation details of these models are given in \nameref{intro other models}. We found that GroundVLP with all models outperforms other zero-shot methods recorded in Table \ref{table:main results rec} , showing its versatility that could be applied to various VLP models effectively. We also report the difference in accuracy between the models obtained on RefCOCO and RefCOCO+ in the last column. As previously mentioned in section \nameref{dataset}, RefCOCO includes more spatial information while RefCOCO+ is composed of more appearance-related queries. Thus, the \textit{Difference} could indicate whether the model is better at position information or appearance attributes. It can be observed that Lxmert is better at recognizing spatial information while the other two end-to-end models are the opposite. We conjecture that the preliminary modeling of objects in the image by the OD module could facilitate the understanding of the visual context for the region-based models and render it position-sensitive.~\cite{yao2022pevl, wang2022position}.

\subsection{Ablation Studies}

The evaluation in this section uses the val split of all datasets and ground-truth category on RefCOCO/+/g datasets if there is no supplementary statement.

\begin{figure}[h]
    \centering
    \includegraphics[width=4.6cm]{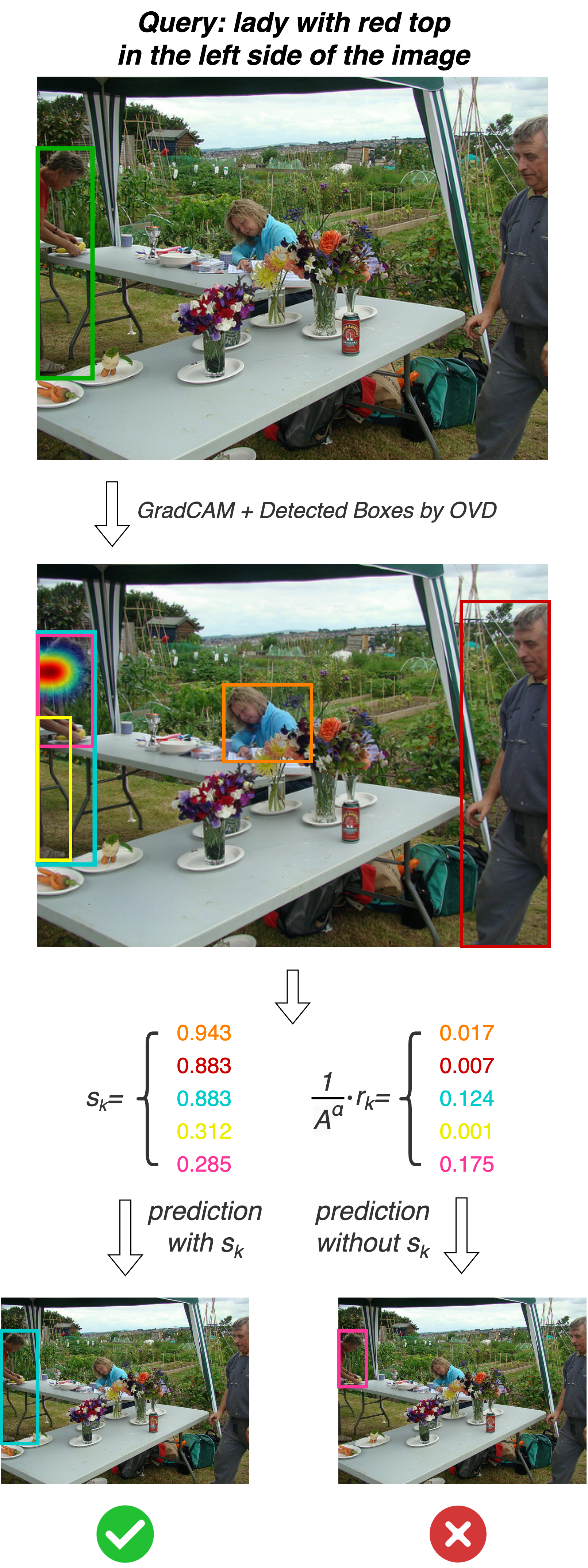}
    \caption{An example of the \textit{weighted grade}. If $s_k$ is not used, GroundVLP would achieve an error prediction. The query is from RefCOCOg val.}
    \label{fig:case5}
\end{figure}

\textbf{Different Assembly of $\theta$ and $s_k$:} Table \ref{table:ablation use sk} investigates the effect of the value of $\theta$ and the usage of the \textit{weighted grade}. It can be observed that a low threshold for OVD ($\theta$ = 0.15) leads to the detection of superfluous boxes,  which will impair the performance if we calculate $r_k$ as grade directly. For this condition, our proposed weighted grade considering both $s_k$ and $r_k$ could effectively mitigate the interference from redundant boxes. A high threshold with the weighted grade ($\theta$ = 0.5 and $s_k$ is used), on the other hand, uplifts the quality of detected boxes but probably excludes the answer box. Thus, the optimal assembly is a relatively low threshold with the \textit{weighted grade}. In Figure \ref{fig:case5}, we illustrate the impact of using the \textit{weighted grade} on the final results. It can be observed that when employing the \textit{weighted grade}, GroundVLP produces the correct answer.

\begin{table}[h!]
  \begin{center}
  \resizebox{\hsize}{!}{
    \setlength\tabcolsep{3pt}
    \begin{tabular}{l c c c c c l l l l} 
    \toprule
      \textbf{Backbone} & & $\theta$ &  &Use $s_{k}$  & & RefCOCO & RefCOCO+ & RefCOCOg & Flickr30k\\
    \midrule
    \multirow{3}{*}{ALBEF}   & \vline& 0.15 &\vline& &\vline & 55.13 & 60.12 & 66.52 & 62.88\\
     & \vline& 0.50 &\vline&\checkmark  &\vline & ${57.31}_{+2.18}$ & ${62.64}_{+2.52} $&${68.24}_{+1.72}$ & ${62.12}_{-0.76}$\\
        & \vline& 0.15  &\vline &\checkmark & \vline & $\textbf{58.22}_{+3.09}$ & $\textbf{63.57}_{+3.45}$&$\textbf{69.63}_{+3.11}$ & $\textbf{63.76}_{+0.88}$\\
      \midrule
      \multirow{3}{*}{VinVL} & \vline& 0.15 &\vline & & \vline&63.82 &67.10 &73.43 & 63.33\\
      & \vline& 0.50 &\vline & \checkmark & \vline& ${64.17}_{+0.35}$& ${68.00}_{+0.90}$ &${72.61}_{-0.82}$ & ${61.01}_{-2.32}$\\
        & \vline& 0.15&\vline & \checkmark&\vline &$\textbf{65.01}_{+1.19}$ & $\textbf{68.87}_{+1.77}$& $\textbf{74.73}_{+1.30}$& $\textbf{63.89}_{+0.56}$\\
    \bottomrule
    \end{tabular}
    }
  \end{center}
  \caption{Ablation study on the value setting of $\theta$ and whether using $s_k$ as the weight of $r_k$.}
  \label{table:ablation use sk}
\end{table}

\begin{table}[h!]
  \begin{center}
  \resizebox{\hsize}{!}{
    \setlength\tabcolsep{3pt}
    \begin{tabular}{l c c c l l l l} 
    \toprule
      \textbf{Backbone} &  &Candidate & & RefCOCO & RefCOCO+ & RefCOCOg & Flickr30k\\
    \midrule
      \multirow{3}{*}{ALBEF} &\vline&\textit{all} &\vline &49.51 &53.80 &55.11 & 55.05\\
      &\vline &\textit{pred} & \vline&$52.58_{+3.07}$ & $56.38_{+2.58}$ &$64.30_{+9.19}$ & $\textbf{63.76}_{+8.71}$\\
       &\vline &\textit{gt} & \vline & $\textbf{58.22}_{+8.71}$ & $\textbf{63.57}_{+9.77}$&$\textbf{69.63}_{+14.52}$ &-\\
      \midrule
         \multirow{3}{*}{VinVL} &\vline & \textit{all}& \vline& 58.39&60.87 & 63.40& 53.23\\
        &\vline &\textit{pred} & \vline& $59.05_{+0.66}$ &$61.80_{+0.93}$ &$69.08_{+5.68}$ & $\textbf{63.89}_{+10.66}$\\
       &\vline &\textit{gt}&\vline & $\textbf{65.01}_{+6.62}$ & $\textbf{68.87}_{+8.00}$& $\textbf{74.73}_{+11.33}$ & - \\
    \bottomrule
    \end{tabular}
    }
  \end{center}
  \caption{Ablation study on the type of candidate boxes. \textit{all} means using all proposals, \textit{pred} means using predicted category, and \textit{gt} means using ground-truth category. }
  \label{table:ablation use all proposals}
\end{table}

\textbf{Type of Candidate Boxes:} Table \ref{table:ablation use all proposals} investigates the influence of different candidate boxes. The performance is enhanced after shrinking the number of candidate boxes by a predetermined category. We observe that the improvements on RefCOCO and RefCOCO+ were not as pronounced as those on RefCOCOg and Flickr30k entities when using the predicted category. Combined with the aforementioned brief about the datasets, it is evident that the extracting for a category is more precise on datasets with concrete and disciplined expressions, such as RefCOCOg and Flickr30k entities that are more practical and common in the real world.

\begin{table}[h!]
  \begin{center}
  \resizebox{\hsize}{!}{
    \setlength\tabcolsep{3pt}
    \begin{tabular}{l c c c l l l l} 
    \toprule
      \textbf{Backbone} &  &Use $vw$  & & RefCOCO & RefCOCO+ & RefCOCOg & Flickr30k\\
    \midrule
     \multirow{2}{*}{ALBEF} &\vline& &\vline & 57.34&63.02 &69.34 & 63.47\\
      &\vline &\checkmark & \vline & $\textbf{58.22}_{+0.88}$ & $\textbf{63.57}_{+0.55}$&$\textbf{69.63}_{+0.29}$ & $\textbf{63.76}_{+0.29}$\\
      \midrule
          \multirow{2}{*}{VinVL}&\vline & & \vline&65.01 &68.28 & 74.53& 63.57\\
       &\vline & \checkmark&\vline & $\textbf{65.01}_{+0.00}$ & $\textbf{68.87}_{+0.59}$& $\textbf{74.73}_{+0.20}$ & $\textbf{63.89}_{+0.32}$\\
        \midrule
         \multirow{2}{*}{PTP} &\vline & & \vline& 55.42 &59.43 &67.89 &-\\
         &\vline & \checkmark&\vline &$\textbf{56.81}_{+1.39}$ &$\textbf{61.17}_{+1.74}$  & $\textbf{68.53}_{+0.64}$& -\\
    \bottomrule
    \end{tabular}
    }
  \end{center}
  \caption{Ablation study on the type of aggregating attention. $vw$ is the simplified spell of visual word attention aggregation. }
  \label{table:ablation use vsiual word}
\end{table}

\textbf{Type of Aggregating Attention:} Table \ref{table:ablation use vsiual word} investigates the improvement of using \textit{visual word attention aggregation.} The compared baseline is implemented by averaging the attention scores across all text tokens (i.e. Figure \ref{fig:three methods} (b)). PTP~\cite{wang2022position} is another VLP we incorporate. It could be observed that our method achieves better performance, especially on PTP, indicating that filtering based on visually recognizable words could facilitate models' text-to-image mapping, which is beneficial for grounding tasks.

\begin{table}[h!]
  \begin{center}
  \resizebox{\hsize}{!}{
    \setlength\tabcolsep{4pt}
    \begin{tabular}{l c c c} 
    \toprule
      \textbf{Model} & RefCOCO & RefCOCO+ & RefCOCOg \\
    \midrule
        Detic & 6.65 & 6.69 & 8.52 \\
        GroundVLP w/ ALBEF & 58.22 & 63.57 & 69.63 \\
    \bottomrule
    \end{tabular}
    }
  \end{center}
  \caption{Comparison of using Detic solely and GroundVLP with ALBEF.}
  \label{table:results ovd}
\end{table}

\textbf{Applying OVD for visual grounding:} In order to figure out whether an OVD could be applied to visual grounding, we exclusively utilized Detic for testing on RefCOCO/+/g. Initially, we fed the ground-truth category into Detic to produce a set of candidate boxes. Subsequently, we input $[query]$ to Detic, taking into account only these candidate boxes when evaluating the similarity score between each proposal and the text embedding of $[query]$. The box with the highest score is the final output. The results in Table \ref{table:results ovd} demonstrate that an OVD can only detect specific categories and struggles to grasp the intricate semantic information of visual grounding without the semantic insight offered by ITM.

\begin{table}[h!]
  \begin{center}
  \resizebox{\hsize}{!}{
    \setlength\tabcolsep{13pt}
    \begin{tabular}{l c c c} 
    \toprule
      \textbf{Model Type} & val & testA & testB \\
    \midrule
        GroundVLP & & & \\
        \quad w/ $ALBEF_{pre-trained}$ & 63.57 & 72.02 & 53.57 \\
        \quad w/ $ALBEF_{fine-tuned}$  & \textbf{68.40} & \textbf{77.07} & \textbf{55.82} \\
    \bottomrule
    \end{tabular}
    }
  \end{center}
  \caption{The comparison of GroundVLP with pre-trained model and fine-tuned model.}
  \label{table:fine-tuned albef}
\end{table}

\subsection{Fine-tuning GroundVLP}
Although we propose GroundVLP as a zero-shot method, it can still be fine-tuned with annotation data to enhance the performance. Using the RefCOCO+ training set, we paired queries with their images as image-text pairs, directly fine-tuned ALBEF using ITM loss and observed performance improvements on RefCOCO+ val and test sets, as shown in Table \ref{table:fine-tuned albef}.

\section{Related Work}

\textbf{Visual Grounding.}\quad The widely used pipelines to resolve visual grounding can be broadly grouped into \textit{two-stage}~\cite{yu2018mattnet, liu2019knowledge,sun2021discriminative} and  \textit{one-stage}~\cite{yang2019fast,liao2020real,deng2021transvg}, where \textit{two-stage} methods exploit a proposal-query matching paradigm while \textit{one-stage} methods generate the answer box with end-to-end. Thanks to the emergence of self-supervised pre-training, the results on visual grounding have been improved substantially by pre-trained models~\cite{chen2020uniter,kamath2021mdetr,li2022grounded}. Additionally, though pre-trained with objectives not related to grounding task, there exist various pre-trained models having a strong capacity for vision-language alignment~\cite{zhang2021vinvl,radford2021learning}.  Thus, the works utilizing their strengths to conduct zero-shot visual grounding were proposed, such as CPT and ReCLIP~\cite{yao2021cpt,subramanian2022reclip}. It is noteworthy that there is another definition of zero-shot different from us, which predicates the objects that are unseen during training and still need to be trained on a grounding dataset~\cite{sadhu2019zero,shi2022improving}. CPT, ReCLIP and us were not trained on any grounding dataset while could carry grounding tasks via triggering the capacity of VLP.

\textbf{GradCAM for Grounding.} GradCAM~\cite{selvaraju2017grad} is proposed to visualize the regions that the model focuses on for a specific output head. When it is used in the VLP models' ITM head, GradCAM could represent a modality mapping from text to image, which is adapted for visual grounding. Therefore, it was employed in REC with a weakly-supervised setting~\cite{li2021align,he2022vlmae} and in robot 3D-navigation~\cite{ha2022semantic} by some works. Different from theirs, our method (1) uses the presented \textit{visual word attention aggregation} to optimize the text-to-image mapping, (2) could generate a heat-map for any VLP models pre-trained with ITM by the approach described in section\ref{Generating a heat-map for VLP}, which is a universal method, and (3) introduce the \textit{weighted grade} to improve the matching between the heat-map and candidate boxes instead of calculating the heat-map values enclosed by boxes directly as other methods.

\section{Conclusion}
We note the ready availability of image-text pairs and object detection data and then present GroundVLP, a zero-shot method for visual grounding via combining the models trained using these datasets. GroundVLP employ GradCAM for a VLP model to identify the image regions, introduce an open-vocabulary object detector to generate the object proposals that belong to a given category, and fuse these two components via the \textit{weighted grade}. Experiments show the state-of-the-art performance of GroundVLP, which outperforms other zero-shot methods and is comparable to some non-VLP-based models. In the future, we plan to refine the method of category prediction and also continue to introduce diverse VLP models and open-vocabulary object detectors for better performance and application of zero-shot grounding models such as embodied agents.

\section{Limitations}
Despite the strong accuracy that GroundVLP achieves, there are still some potential limitations. GroundVLP may inadvertently inherit biases or errors presented in those foundational models, such as the results shown in Table \ref{table:results size variation}. However, we note that both VLP and OVD serve as plug-and-play modules in our implementation. This modular design means that GroundVLP stands to benefit from advancements in both of these areas. Should there be a more robust or improved foundational model in the future, it can be seamlessly integrated into our framework to replace any prior model exhibiting errors or biases. Furthermore, as shown in Table~\ref{table:fine-tuned albef}, GroundVLP achieves a performance improvement after fine-tuning its VLP backbone, showing that the employed foundation model could be fine-tuned to effectively alleviate these deficiencies.

\section{Acknowledgements}
This research is supported by National Key R\&D Program of China under grant (2022YFF0902600) and Key R\&D Program of Zhejiang under grant (2023C01048).

\bibliography{aaai24}

\begin{thebibliography}{52}
\providecommand{\natexlab}[1]{#1}

\bibitem[{Anderson et~al.(2018)Anderson, He, Buehler, Teney, Johnson, Gould, and Zhang}]{anderson2018bottom}
Anderson, P.; He, X.; Buehler, C.; Teney, D.; Johnson, M.; Gould, S.; and Zhang, L. 2018.
\newblock Bottom-up and top-down attention for image captioning and visual question answering.
\newblock In \emph{Proceedings of the IEEE conference on computer vision and pattern recognition}, 6077--6086.

\bibitem[{Changpinyo et~al.(2021)Changpinyo, Sharma, Ding, and Soricut}]{changpinyo2021conceptual}
Changpinyo, S.; Sharma, P.; Ding, N.; and Soricut, R. 2021.
\newblock Conceptual 12m: Pushing web-scale image-text pre-training to recognize long-tail visual concepts.
\newblock In \emph{Proceedings of the IEEE/CVF Conference on Computer Vision and Pattern Recognition}, 3558--3568.

\bibitem[{Chen et~al.(2020{\natexlab{a}})Chen, Li, Yu, El~Kholy, Ahmed, Gan, Cheng, and Liu}]{chen2020uniter}
Chen, Y.-C.; Li, L.; Yu, L.; El~Kholy, A.; Ahmed, F.; Gan, Z.; Cheng, Y.; and Liu, J. 2020{\natexlab{a}}.
\newblock Uniter: Universal image-text representation learning.
\newblock In \emph{Computer Vision--ECCV 2020: 16th European Conference, Glasgow, UK, August 23--28, 2020, Proceedings, Part XXX}, 104--120. Springer.

\bibitem[{Chen et~al.(2020{\natexlab{b}})Chen, Wang, Ma, Wong, and Wu}]{chen2020cops}
Chen, Z.; Wang, P.; Ma, L.; Wong, K.-Y.~K.; and Wu, Q. 2020{\natexlab{b}}.
\newblock Cops-ref: A new dataset and task on compositional referring expression comprehension.
\newblock In \emph{Proceedings of the IEEE/CVF Conference on Computer Vision and Pattern Recognition}, 10086--10095.

\bibitem[{Deng et~al.(2009)Deng, Dong, Socher, Li, Li, and Fei-Fei}]{deng2009imagenet}
Deng, J.; Dong, W.; Socher, R.; Li, L.-J.; Li, K.; and Fei-Fei, L. 2009.
\newblock Imagenet: A large-scale hierarchical image database.
\newblock In \emph{2009 IEEE conference on computer vision and pattern recognition}, 248--255. Ieee.

\bibitem[{Deng et~al.(2021)Deng, Yang, Chen, Zhou, and Li}]{deng2021transvg}
Deng, J.; Yang, Z.; Chen, T.; Zhou, W.; and Li, H. 2021.
\newblock Transvg: End-to-end visual grounding with transformers.
\newblock In \emph{Proceedings of the IEEE/CVF International Conference on Computer Vision}, 1769--1779.

\bibitem[{Dosovitskiy et~al.(2020)Dosovitskiy, Beyer, Kolesnikov, Weissenborn, Zhai, Unterthiner, Dehghani, Minderer, Heigold, Gelly et~al.}]{dosovitskiy2020image}
Dosovitskiy, A.; Beyer, L.; Kolesnikov, A.; Weissenborn, D.; Zhai, X.; Unterthiner, T.; Dehghani, M.; Minderer, M.; Heigold, G.; Gelly, S.; et~al. 2020.
\newblock An Image is Worth 16x16 Words: Transformers for Image Recognition at Scale.
\newblock In \emph{International Conference on Learning Representations}.

\bibitem[{Gu et~al.(2021)Gu, Lin, Kuo, and Cui}]{gu2021open}
Gu, X.; Lin, T.-Y.; Kuo, W.; and Cui, Y. 2021.
\newblock Open-Vocabulary Detection via Vision and Language Knowledge Distillation.
\newblock \emph{arXiv preprint arXiv:2104.13921}.

\bibitem[{Gupta, Dollar, and Girshick(2019)}]{gupta2019lvis}
Gupta, A.; Dollar, P.; and Girshick, R. 2019.
\newblock Lvis: A dataset for large vocabulary instance segmentation.
\newblock In \emph{Proceedings of the IEEE/CVF conference on computer vision and pattern recognition}, 5356--5364.

\bibitem[{Ha and Song(2022)}]{ha2022semantic}
Ha, H.; and Song, S. 2022.
\newblock Semantic abstraction: Open-world 3d scene understanding from 2d vision-language models.
\newblock In \emph{Conference on Robot Learning}.

\bibitem[{He et~al.(2022)He, Guo, Dai, Qiao, Wu, Shu, and Ren}]{he2022vlmae}
He, S.; Guo, T.; Dai, T.; Qiao, R.; Wu, C.; Shu, X.; and Ren, B. 2022.
\newblock VLMAE: Vision-Language Masked Autoencoder.
\newblock \emph{arXiv preprint arXiv:2208.09374}.

\bibitem[{Honnibal and Johnson(2015)}]{honnibal2015improved}
Honnibal, M.; and Johnson, M. 2015.
\newblock An improved non-monotonic transition system for dependency parsing.
\newblock In \emph{Proceedings of the 2015 conference on empirical methods in natural language processing}, 1373--1378.

\bibitem[{Kamath et~al.(2021)Kamath, Singh, LeCun, Synnaeve, Misra, and Carion}]{kamath2021mdetr}
Kamath, A.; Singh, M.; LeCun, Y.; Synnaeve, G.; Misra, I.; and Carion, N. 2021.
\newblock Mdetr-modulated detection for end-to-end multi-modal understanding.
\newblock In \emph{Proceedings of the IEEE/CVF International Conference on Computer Vision}, 1780--1790.

\bibitem[{Kim, Jun, and Zhang(2018)}]{kim2018bilinear}
Kim, J.-H.; Jun, J.; and Zhang, B.-T. 2018.
\newblock Bilinear attention networks.
\newblock \emph{Advances in neural information processing systems}, 31.

\bibitem[{Kim, Son, and Kim(2021)}]{kim2021vilt}
Kim, W.; Son, B.; and Kim, I. 2021.
\newblock Vilt: Vision-and-language transformer without convolution or region supervision.
\newblock In \emph{International Conference on Machine Learning}, 5583--5594. PMLR.

\bibitem[{Krishna et~al.(2017)Krishna, Zhu, Groth, Johnson, Hata, Kravitz, Chen, Kalantidis, Li, Shamma et~al.}]{krishna2017visual}
Krishna, R.; Zhu, Y.; Groth, O.; Johnson, J.; Hata, K.; Kravitz, J.; Chen, S.; Kalantidis, Y.; Li, L.-J.; Shamma, D.~A.; et~al. 2017.
\newblock Visual genome: Connecting language and vision using crowdsourced dense image annotations.
\newblock \emph{International journal of computer vision}, 123: 32--73.

\bibitem[{Kuznetsova et~al.(2020)Kuznetsova, Rom, Alldrin, Uijlings, Krasin, Pont-Tuset, Kamali, Popov, Malloci, Kolesnikov et~al.}]{kuznetsova2020open}
Kuznetsova, A.; Rom, H.; Alldrin, N.; Uijlings, J.; Krasin, I.; Pont-Tuset, J.; Kamali, S.; Popov, S.; Malloci, M.; Kolesnikov, A.; et~al. 2020.
\newblock The open images dataset v4: Unified image classification, object detection, and visual relationship detection at scale.
\newblock \emph{International Journal of Computer Vision}, 128(7): 1956--1981.

\bibitem[{Li et~al.(2022{\natexlab{a}})Li, Li, Xiong, and Hoi}]{li2022blip}
Li, J.; Li, D.; Xiong, C.; and Hoi, S. 2022{\natexlab{a}}.
\newblock Blip: Bootstrapping language-image pre-training for unified vision-language understanding and generation.
\newblock In \emph{International Conference on Machine Learning}, 12888--12900. PMLR.

\bibitem[{Li et~al.(2021)Li, Selvaraju, Gotmare, Joty, Xiong, and Hoi}]{li2021align}
Li, J.; Selvaraju, R.; Gotmare, A.; Joty, S.; Xiong, C.; and Hoi, S. C.~H. 2021.
\newblock Align before fuse: Vision and language representation learning with momentum distillation.
\newblock \emph{Advances in neural information processing systems}, 34: 9694--9705.

\bibitem[{Li et~al.(2022{\natexlab{b}})Li, Zhang, Zhang, Yang, Li, Zhong, Wang, Yuan, Zhang, Hwang et~al.}]{li2022grounded}
Li, L.~H.; Zhang, P.; Zhang, H.; Yang, J.; Li, C.; Zhong, Y.; Wang, L.; Yuan, L.; Zhang, L.; Hwang, J.-N.; et~al. 2022{\natexlab{b}}.
\newblock Grounded language-image pre-training.
\newblock In \emph{Proceedings of the IEEE/CVF Conference on Computer Vision and Pattern Recognition}, 10965--10975.

\bibitem[{Liao et~al.(2020)Liao, Liu, Li, Wang, Chen, Qian, and Li}]{liao2020real}
Liao, Y.; Liu, S.; Li, G.; Wang, F.; Chen, Y.; Qian, C.; and Li, B. 2020.
\newblock A real-time cross-modality correlation filtering method for referring expression comprehension.
\newblock In \emph{Proceedings of the IEEE/CVF Conference on Computer Vision and Pattern Recognition}, 10880--10889.

\bibitem[{Lin et~al.(2014)Lin, Maire, Belongie, Hays, Perona, Ramanan, Doll{\'a}r, and Zitnick}]{lin2014microsoft}
Lin, T.-Y.; Maire, M.; Belongie, S.; Hays, J.; Perona, P.; Ramanan, D.; Doll{\'a}r, P.; and Zitnick, C.~L. 2014.
\newblock Microsoft coco: Common objects in context.
\newblock In \emph{Computer Vision--ECCV 2014: 13th European Conference, Zurich, Switzerland, September 6-12, 2014, Proceedings, Part V 13}, 740--755. Springer.

\bibitem[{Liu et~al.(2019{\natexlab{a}})Liu, Zhang, Wu, and Zha}]{liu2019learning}
Liu, D.; Zhang, H.; Wu, F.; and Zha, Z.-J. 2019{\natexlab{a}}.
\newblock Learning to assemble neural module tree networks for visual grounding.
\newblock In \emph{Proceedings of the IEEE/CVF International Conference on Computer Vision}, 4673--4682.

\bibitem[{Liu et~al.(2019{\natexlab{b}})Liu, Liu, Bai, and Yuille}]{liu2019clevr}
Liu, R.; Liu, C.; Bai, Y.; and Yuille, A.~L. 2019{\natexlab{b}}.
\newblock Clevr-ref+: Diagnosing visual reasoning with referring expressions.
\newblock In \emph{Proceedings of the IEEE/CVF conference on computer vision and pattern recognition}, 4185--4194.

\bibitem[{Liu et~al.(2019{\natexlab{c}})Liu, Li, Wang, Zha, Su, and Huang}]{liu2019knowledge}
Liu, X.; Li, L.; Wang, S.; Zha, Z.-J.; Su, L.; and Huang, Q. 2019{\natexlab{c}}.
\newblock Knowledge-guided pairwise reconstruction network for weakly supervised referring expression grounding.
\newblock In \emph{Proceedings of the 27th ACM International Conference on Multimedia}, 539--547.

\bibitem[{Mao et~al.(2016)Mao, Huang, Toshev, Camburu, Yuille, and Murphy}]{mao2016generation}
Mao, J.; Huang, J.; Toshev, A.; Camburu, O.; Yuille, A.~L.; and Murphy, K. 2016.
\newblock Generation and comprehension of unambiguous object descriptions.
\newblock In \emph{Proceedings of the IEEE conference on computer vision and pattern recognition}, 11--20.

\bibitem[{Ordonez, Kulkarni, and Berg(2011)}]{ordonez2011im2text}
Ordonez, V.; Kulkarni, G.; and Berg, T. 2011.
\newblock Im2text: Describing images using 1 million captioned photographs.
\newblock \emph{Advances in neural information processing systems}, 24.

\bibitem[{Plummer et~al.(2015)Plummer, Wang, Cervantes, Caicedo, Hockenmaier, and Lazebnik}]{plummer2015flickr30k}
Plummer, B.~A.; Wang, L.; Cervantes, C.~M.; Caicedo, J.~C.; Hockenmaier, J.; and Lazebnik, S. 2015.
\newblock Flickr30k entities: Collecting region-to-phrase correspondences for richer image-to-sentence models.
\newblock In \emph{Proceedings of the IEEE international conference on computer vision}, 2641--2649.

\bibitem[{Qi et~al.(2020)Qi, Zhang, Zhang, Bolton, and Manning}]{qi2020stanza}
Qi, P.; Zhang, Y.; Zhang, Y.; Bolton, J.; and Manning, C.~D. 2020.
\newblock Stanza: A Python Natural Language Processing Toolkit for Many Human Languages.
\newblock In \emph{ACL (demo)}.

\bibitem[{Radford et~al.(2021)Radford, Kim, Hallacy, Ramesh, Goh, Agarwal, Sastry, Askell, Mishkin, Clark et~al.}]{radford2021learning}
Radford, A.; Kim, J.~W.; Hallacy, C.; Ramesh, A.; Goh, G.; Agarwal, S.; Sastry, G.; Askell, A.; Mishkin, P.; Clark, J.; et~al. 2021.
\newblock Learning transferable visual models from natural language supervision.
\newblock In \emph{International Conference on Machine Learning}, 8748--8763. PMLR.

\bibitem[{Sadhu, Chen, and Nevatia(2019)}]{sadhu2019zero}
Sadhu, A.; Chen, K.; and Nevatia, R. 2019.
\newblock Zero-shot grounding of objects from natural language queries.
\newblock In \emph{Proceedings of the IEEE/CVF International Conference on Computer Vision}, 4694--4703.

\bibitem[{Schuhmann et~al.(2022)Schuhmann, Beaumont, Vencu, Gordon, Wightman, Cherti, Coombes, Katta, Mullis, Wortsman et~al.}]{schuhmann2022laion}
Schuhmann, C.; Beaumont, R.; Vencu, R.; Gordon, C.; Wightman, R.; Cherti, M.; Coombes, T.; Katta, A.; Mullis, C.; Wortsman, M.; et~al. 2022.
\newblock Laion-5b: An open large-scale dataset for training next generation image-text models.
\newblock \emph{Advances in Neural Information Processing Systems}, 35: 25278--25294.

\bibitem[{Schuhmann et~al.(2021)Schuhmann, Vencu, Beaumont, Kaczmarczyk, Mullis, Katta, Coombes, Jitsev, and Komatsuzaki}]{schuhmann2021laion}
Schuhmann, C.; Vencu, R.; Beaumont, R.; Kaczmarczyk, R.; Mullis, C.; Katta, A.; Coombes, T.; Jitsev, J.; and Komatsuzaki, A. 2021.
\newblock Laion-400m: Open dataset of clip-filtered 400 million image-text pairs.
\newblock \emph{arXiv preprint arXiv:2111.02114}.

\bibitem[{Selvaraju et~al.(2017)Selvaraju, Cogswell, Das, Vedantam, Parikh, and Batra}]{selvaraju2017grad}
Selvaraju, R.~R.; Cogswell, M.; Das, A.; Vedantam, R.; Parikh, D.; and Batra, D. 2017.
\newblock Grad-cam: Visual explanations from deep networks via gradient-based localization.
\newblock In \emph{Proceedings of the IEEE international conference on computer vision}, 618--626.

\bibitem[{Shi et~al.(2022)Shi, Shen, Jin, and Zhu}]{shi2022improving}
Shi, Z.; Shen, Y.; Jin, H.; and Zhu, X. 2022.
\newblock Improving Zero-Shot Phrase Grounding via Reasoning on External Knowledge and Spatial Relations.
\newblock In \emph{Proceedings of the AAAI Conference on Artificial Intelligence}, volume~36, 2253--2261.

\bibitem[{Subramanian et~al.(2022)Subramanian, Merrill, Darrell, Gardner, Singh, and Rohrbach}]{subramanian2022reclip}
Subramanian, S.; Merrill, W.; Darrell, T.; Gardner, M.; Singh, S.; and Rohrbach, A. 2022.
\newblock ReCLIP: A Strong Zero-Shot Baseline for Referring Expression Comprehension.
\newblock In \emph{Proceedings of the 60th Annual Meeting of the Association for Computational Linguistics (Volume 1: Long Papers)}, 5198--5215.

\bibitem[{Sun et~al.(2021)Sun, Xiao, Lim, Liu, and Goulermas}]{sun2021discriminative}
Sun, M.; Xiao, J.; Lim, E.~G.; Liu, S.; and Goulermas, J.~Y. 2021.
\newblock Discriminative triad matching and reconstruction for weakly referring expression grounding.
\newblock \emph{IEEE transactions on pattern analysis and machine intelligence}, 43(11): 4189--4195.

\bibitem[{Tan and Bansal(2019)}]{tan2019lxmert}
Tan, H.; and Bansal, M. 2019.
\newblock LXMERT: Learning Cross-Modality Encoder Representations from Transformers.
\newblock In \emph{Proceedings of the 2019 Conference on Empirical Methods in Natural Language Processing and the 9th International Joint Conference on Natural Language Processing (EMNLP-IJCNLP)}, 5100--5111.

\bibitem[{Vaswani et~al.(2017)Vaswani, Shazeer, Parmar, Uszkoreit, Jones, Gomez, Kaiser, and Polosukhin}]{vaswani2017attention}
Vaswani, A.; Shazeer, N.; Parmar, N.; Uszkoreit, J.; Jones, L.; Gomez, A.~N.; Kaiser, {\L}.; and Polosukhin, I. 2017.
\newblock Attention is all you need.
\newblock \emph{Advances in neural information processing systems}, 30.

\bibitem[{Wang et~al.(2022)Wang, Zhou, Shou, and Yan}]{wang2022position}
Wang, A.~J.; Zhou, P.; Shou, M.~Z.; and Yan, S. 2022.
\newblock Position-guided Text Prompt for Vision-Language Pre-training.
\newblock \emph{arXiv preprint arXiv:2212.09737}.

\bibitem[{Yan et~al.(2023)Yan, Jiang, Wu, Wang, Luo, Yuan, and Lu}]{yan2023universal}
Yan, B.; Jiang, Y.; Wu, J.; Wang, D.; Luo, P.; Yuan, Z.; and Lu, H. 2023.
\newblock Universal instance perception as object discovery and retrieval.
\newblock In \emph{Proceedings of the IEEE/CVF Conference on Computer Vision and Pattern Recognition}, 15325--15336.

\bibitem[{Yang et~al.(2022)Yang, Duan, Tran, Xu, Chanda, Chen, Zeng, Chilimbi, and Huang}]{yang2022vision}
Yang, J.; Duan, J.; Tran, S.; Xu, Y.; Chanda, S.; Chen, L.; Zeng, B.; Chilimbi, T.; and Huang, J. 2022.
\newblock Vision-language pre-training with triple contrastive learning.
\newblock In \emph{Proceedings of the IEEE/CVF Conference on Computer Vision and Pattern Recognition}, 15671--15680.

\bibitem[{Yang et~al.(2019)Yang, Gong, Wang, Huang, Yu, and Luo}]{yang2019fast}
Yang, Z.; Gong, B.; Wang, L.; Huang, W.; Yu, D.; and Luo, J. 2019.
\newblock A fast and accurate one-stage approach to visual grounding.
\newblock In \emph{Proceedings of the IEEE/CVF International Conference on Computer Vision}, 4683--4693.

\bibitem[{Yao et~al.(2022)Yao, Chen, Zhang, Ji, Liu, Chua, and Sun}]{yao2022pevl}
Yao, Y.; Chen, Q.; Zhang, A.; Ji, W.; Liu, Z.; Chua, T.-S.; and Sun, M. 2022.
\newblock PEVL: Position-enhanced pre-training and prompt tuning for vision-language models.
\newblock \emph{arXiv preprint arXiv:2205.11169}.

\bibitem[{Yao et~al.(2021)Yao, Zhang, Zhang, Liu, Chua, and Sun}]{yao2021cpt}
Yao, Y.; Zhang, A.; Zhang, Z.; Liu, Z.; Chua, T.-S.; and Sun, M. 2021.
\newblock Cpt: Colorful prompt tuning for pre-trained vision-language models.
\newblock \emph{arXiv preprint arXiv:2109.11797}.

\bibitem[{Yu et~al.(2018)Yu, Lin, Shen, Yang, Lu, Bansal, and Berg}]{yu2018mattnet}
Yu, L.; Lin, Z.; Shen, X.; Yang, J.; Lu, X.; Bansal, M.; and Berg, T.~L. 2018.
\newblock Mattnet: Modular attention network for referring expression comprehension.
\newblock In \emph{Proceedings of the IEEE conference on computer vision and pattern recognition}, 1307--1315.

\bibitem[{Yu et~al.(2016)Yu, Poirson, Yang, Berg, and Berg}]{yu2016modeling}
Yu, L.; Poirson, P.; Yang, S.; Berg, A.~C.; and Berg, T.~L. 2016.
\newblock Modeling context in referring expressions.
\newblock In \emph{Computer Vision--ECCV 2016: 14th European Conference, Amsterdam, The Netherlands, October 11-14, 2016, Proceedings, Part II 14}, 69--85. Springer.

\bibitem[{Zareian et~al.(2021)Zareian, Rosa, Hu, and Chang}]{zareian2021open}
Zareian, A.; Rosa, K.~D.; Hu, D.~H.; and Chang, S.-F. 2021.
\newblock Open-vocabulary object detection using captions.
\newblock In \emph{Proceedings of the IEEE/CVF Conference on Computer Vision and Pattern Recognition}, 14393--14402.

\bibitem[{Zhang et~al.(2021)Zhang, Li, Hu, Yang, Zhang, Wang, Choi, and Gao}]{zhang2021vinvl}
Zhang, P.; Li, X.; Hu, X.; Yang, J.; Zhang, L.; Wang, L.; Choi, Y.; and Gao, J. 2021.
\newblock Vinvl: Revisiting visual representations in vision-language models.
\newblock In \emph{Proceedings of the IEEE/CVF Conference on Computer Vision and Pattern Recognition}, 5579--5588.

\bibitem[{Zhao et~al.(2022{\natexlab{a}})Zhao, Liu, Lu, and Lee}]{zhao2022omdet}
Zhao, T.; Liu, P.; Lu, X.; and Lee, K. 2022{\natexlab{a}}.
\newblock OmDet: Language-Aware Object Detection with Large-scale Vision-Language Multi-dataset Pre-training.
\newblock \emph{arXiv preprint arXiv:2209.05946}.

\bibitem[{Zhao et~al.(2022{\natexlab{b}})Zhao, Zhang, Zhu, Shen, Lee, Lu, and Yin}]{zhao2022vl}
Zhao, T.; Zhang, T.; Zhu, M.; Shen, H.; Lee, K.; Lu, X.; and Yin, J. 2022{\natexlab{b}}.
\newblock VL-CheckList: Evaluating Pre-trained Vision-Language Models with Objects, Attributes and Relations.
\newblock \emph{arXiv preprint arXiv:2207.00221}.

\bibitem[{Zhou et~al.(2022)Zhou, Girdhar, Joulin, Kr{\"a}henb{\"u}hl, and Misra}]{zhou2022detecting}
Zhou, X.; Girdhar, R.; Joulin, A.; Kr{\"a}henb{\"u}hl, P.; and Misra, I. 2022.
\newblock Detecting twenty-thousand classes using image-level supervision.
\newblock In \emph{Computer Vision--ECCV 2022: 17th European Conference, Tel Aviv, Israel, October 23--27, 2022, Proceedings, Part IX}, 350--368. Springer.

\end{thebibliography}

\appendix

\section{Descriptions and Implementation Details of Other Pre-trained Models} \label{intro other models}

\noindent\textbf{TCL}~\cite{yang2022vision}, a two-stream end-to-end model, is an enhanced version of ALBEF, which introduces three contrasting modules: Cross-modal Alignment (CMA), Intra-modal Contrastive (IMC), and Local Mutual Information Maximization (LMI). These modules are designed to maximize the mutual information between matching images and text and maximize global mutual information. Conforming to ALBEF, we use the $3^{rd}$ layer of the cross-modality fusion encoder for GradCAM. We adopt its TCL-4M checkpoint and the input prompt is the same as ALBEF described in the main text.

\noindent\textbf{PTP}~\cite{wang2022position} exploits a position-guided text prompt for VLP models to embed the positional information during training. It has two versions introducing ViLT~\cite{kim2021vilt} and BLIP~\cite{li2022blip} as the backbone, respectively. We choose the one introducing BLIP, which is a two-stream end-to-end model, and use GradCAM in the $8^{th}$ layer of its cross-modality fusion encoder. We adopt its PTP-BLIP-4M checkpoint and the input prompt is the same as ALBEF described in the main text.

\noindent\textbf{Lxmert}~\cite{tan2019lxmert} is a two-stream region-based model, depending on a widely used bottom-up and top-down object detector~\cite{anderson2018bottom} to generate visual features. The model is pretrained with ITM and other three objectives. Generally, the co-attention module of cross-attention encoder in two-stream models collects \textbf{K}, \textbf{V} from image modality and \textbf{Q} from text modality, leading the shape of \textbf{A} calculated by $ \textbf{Q}\cdot \textbf{K}^\top$ is $\textit{T}\times\textit{I}$, whereas Lxmert has two co-attention modules, one of which collects \textbf{K}, \textbf{V} from image modality and \textbf{Q} from text as usual and another is the opposite. We use GradCAM in the former one so that the attention map can represent the influence of image tokens on each text token as in other two-stream models. Furthermore, we set \textit{m} as 5 for Lxmert and use the $3^{rd}$ layer of its cross-modality fusion encoder. We adopt its Lxmert-20epochs checkpoint and the input prompt is the same as ALBEF described in the main text.

\section{Robustness against the Variation in Size} \label{Variation in Size}

Following the example of VL-CheckList~\cite{zhao2022vl}, we divide RefCOCO/+/g into \textit{small} and \textit{large} split by the area of referring target. The result in \ref{table:results size variation} shows a similar conclusion to VL-CheckList that VLP models tend to keep a watch on large objects, showing its weak robustness against the variation in size.

\begin{table}[h!]
  \begin{center}
  \resizebox{\hsize}{!}{
    \setlength\tabcolsep{4pt}
    \begin{tabular}{l c c c c c c c c} 
    \toprule
       & \multicolumn{2}{c}{RefCOCO} & & \multicolumn{2}{c}{RefCOCO+} & & \multicolumn{2}{c}{RefCOCOg}\\
      \textbf{Model} & small & large & & small & large & & small & large  \\
    \midrule
      \textbf{GroundVLP} & & & \vline & & & \vline & & \\
      \quad w/ ALBEF & 40.53 & 78.76 & \vline & 44.65 & 82.24 & \vline & 58.51 & 87.07\\ 
      \quad w/ VinVL & 52.38 & 79.16 & \vline & 53.00 & 85.01 & \vline & 67.42 & 88.89\\ 
    \bottomrule
    \end{tabular}
    }
  \end{center}
  \caption{Accuracy (\%) of GroundVLP on different size splits, \textit{small} refers to the ratio of the area of referring target to the full image is less than \textit{0.1}, while \textit{large} means the ratio is greater than \textit{0.4}. All datasets in the table indicate their validation split.}
  \label{table:results size variation}
\end{table}

\section{Case Study}\label{case study}
\subsection{GroundVLP with different VLP models}
\begin{figure*}[h]
    \centering
    \includegraphics[width=13cm]{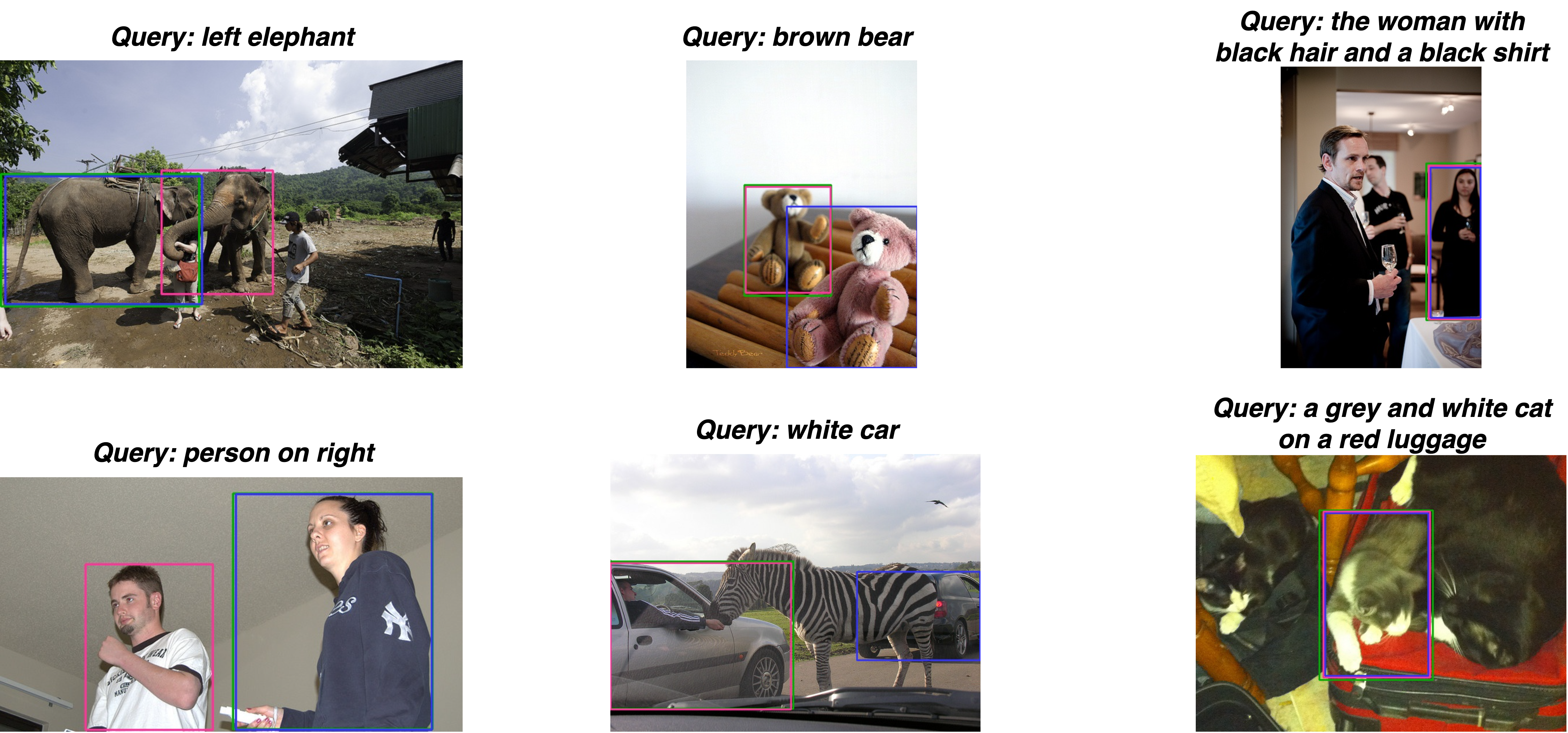}
    \caption{Case study on the difference of the predictions from ALBEF\,(\textcolor[RGB]{255,51,153}{pink}) and VinVL\,(\textcolor[RGB]{51,51,255}{blue}). The ground-truth bounding box is colored with \textcolor[RGB]{0,176,0}{green}. We show some results of GroundVLP: cases where only VinVL predicts correctly\,(left), cases where only ALBEF predicts correctly\,(middle), and cases where both ALBEF and VinVL predict correctly\,(right). All results are obtained by using ground-truth category. The queries in left, middle and right are from RefCOCO, RefCOCO+ and RefCOCOg, respectively.}
    \label{fig:case}
\end{figure*}

We visualize the predictions of GroundVLP with different VLP models in Figure \ref{fig:case}. VinVL, a region-based model, excels in queries with position information compared to ALBEF, which belongs to end-to-end and perform well in queries depicted by appearance attributes. We observe that GroundVLP with both models achieves superior performance for long and concrete queries, indicating that our method leverages the pre-training capacity of VLP models to facilitate the text-region alignment when there is more semantic information provided.

\subsection{Noise to Disturb the Category Extraction}\label{noise}
\begin{figure*}[h]
    \centering
    \includegraphics[width=14cm]{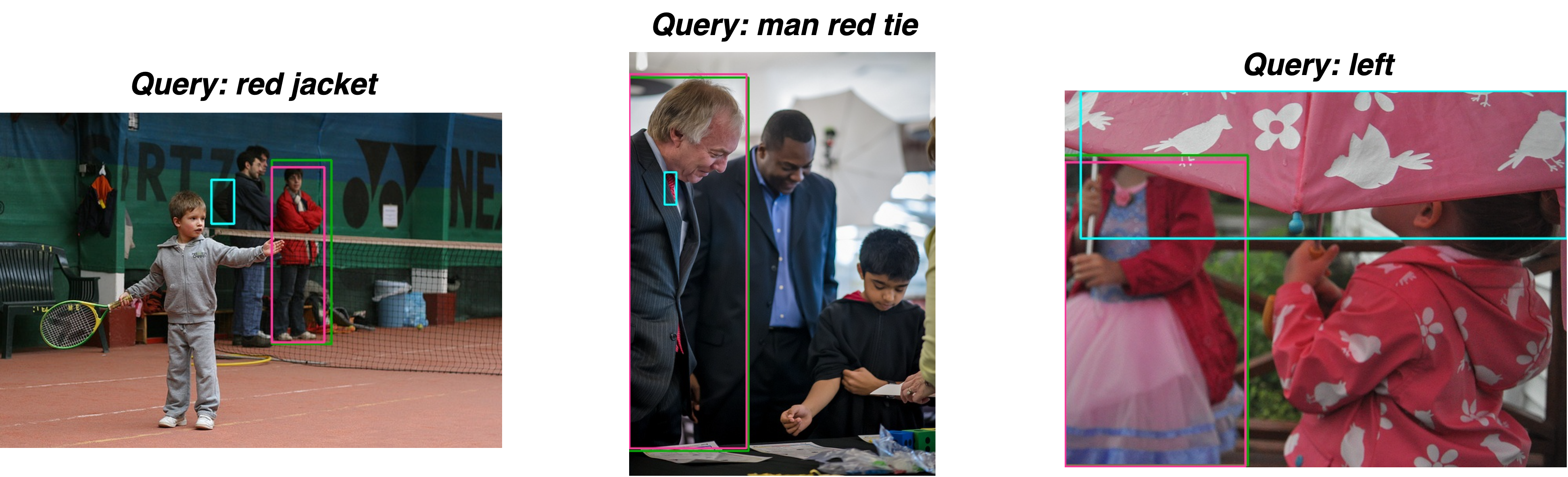}
    \caption{Case study on the textual noise from datasets. We show the results of GroundVLP with ALBEF using ground-truth category\,(\textcolor[RGB]{255,51,153}{pink}) and predicted category\,(\textcolor[RGB]{0,200,200}{cyan}). The ground-truth bounding box is colored with \textcolor[RGB]{0,176,0}{green}. The three queries are from RefCOCO+, RefCOCO+ and RefCOCO, respectively.}
    \label{fig:case2}
\end{figure*}

\begin{figure*}[h]
    \centering
    \includegraphics[width=14cm]{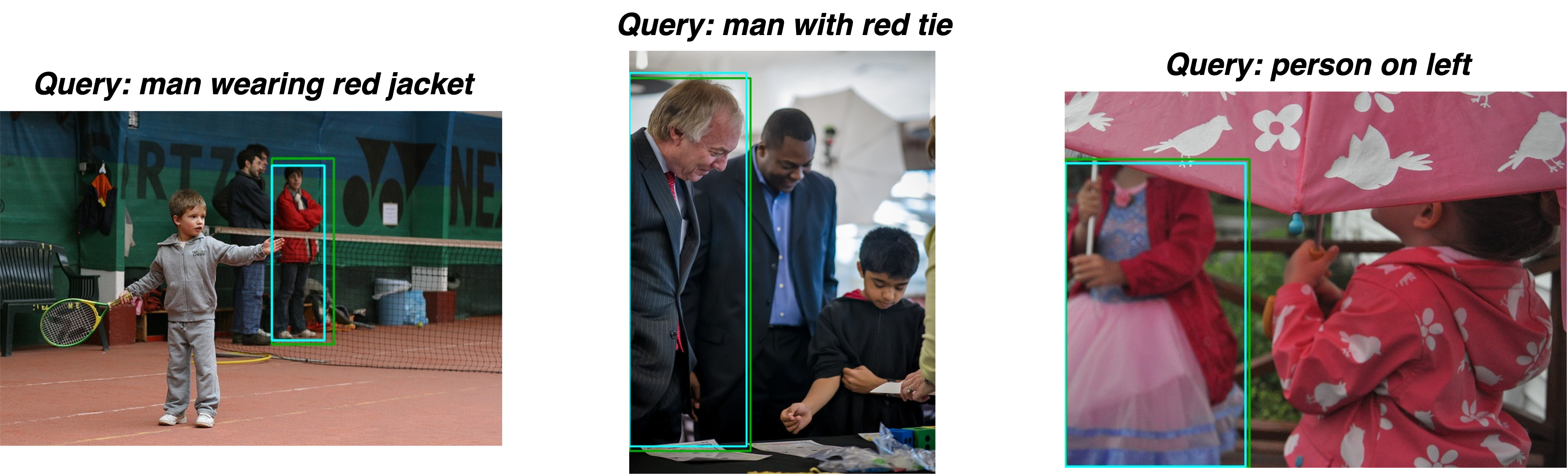}
    \caption{The predictions of the corrected queries. We show the results of GroundVLP with ALBEF using predicted category\,(\textcolor[RGB]{0,200,200}{cyan}). The ground-truth bounding box is colored with \textcolor[RGB]{0,176,0}{green}. The images are the same as that in Figure \ref{fig:case2}. }
    \label{fig:case3}
\end{figure*}

We visualize the predictions of GroundVLP using different types of category in Figure \ref{fig:case2}. In the main text, we discuss three types of noise found in RefCOCO/+/g datasets, each of which is provided with an example here:\,(1)\,Unclear referring targets: ``\textit{red jacket}" refers to a person wearing red jacket, while our used NLP toolbox would extract ``\textit{jacket}" as the predicted category for this unclear query and it will be mapped into the COCO class ``\textit{backpack}". Thus the open-vocabulary object detector generates the candidate boxes for \textit{backpack} instead of \textit{person}, bringing about the mistake. (2)\,Undisciplined grammar: ``\textit{man red tie}" is the undisciplined expression of ``\textit{man with red tie}". It causes the NLP toolbox to identify ``tie" as the target unit rather than ``man" so that the open-vocabulary detector cannot generate candidate boxes by rule and line. (3)\,No target in query: the query ``\textit{left}" includes no target unit, causing all proposals to be used as candidate boxes and resulting in an incorrect prediction. We correct some deficiencies in these queries, and GroundVLP achieves the correct results, as shown in Figure \ref{fig:case3}.

\subsection{Phrase Grounding}
\begin{figure}[h]
    \centering
    \includegraphics[width=8cm]{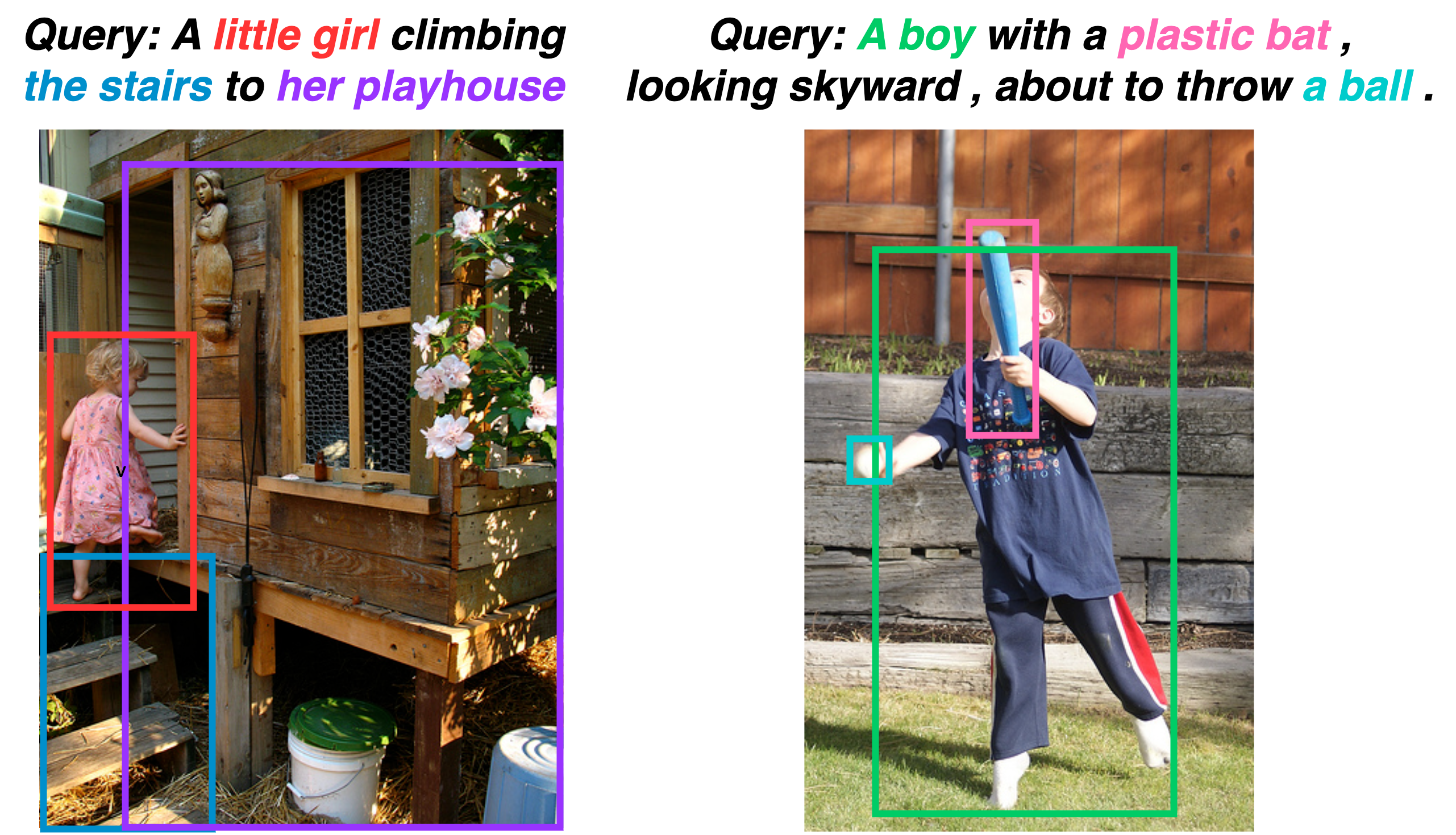}
    \caption{The predictions of the GroundVLP with ALBEF for phrase grounding. The queries in left and right are from Flickr30k entities val and test, respectively.}
    \label{fig:case4}
\end{figure}

We further show the visualization of phrase grounding\,(Figure \ref{fig:case4}).

\section{Introduction of the Input Format to VinVL}
\label{appendix:vinvl input}
$\mathcal{I}, \mathcal{T}$ is defined as the input image and text. The input format of VinVL is a triple tuple $\left\{ \textbf{\textit{w}}\,\ \textbf{\textit{q}}\,\ \textbf{\textit{v}} \right\}$ and could be interpreted by two ways: (1) \textbf{\textit{w}} denotes a caption for $\mathcal{I}$, \textbf{\textit{q}} denotes the object labels detected by its OD module, and \textbf{\textit{v}} is the visual features obtained by the OD. VinVL could predict whether \textbf{\textit{w-v}} is a matched text-image pair and output the logical score through the ITM head. (2) \textbf{\textit{w}} denotes a question about $\mathcal{I}$ , \textbf{\textit{q}} denotes the answer for \textbf{\textit{w}}, and \textbf{\textit{v}} is the visual features obtained by the OD. VinVL could predict whether \textbf{\textit{w-q}} is a matched question-answer pair and output the logical score through the ITM head. We name them as ITM-resemble and VQA-resemble, respectively. For more details on why this input format was chosen and further information about VinVL, please refer to the original paper~\cite{zhang2021vinvl}.

\section{Example of CPT-adapted}\label{appendix:cpt-adapted}
\begin{figure}[h]
    \centering
    \includegraphics[width=8cm]{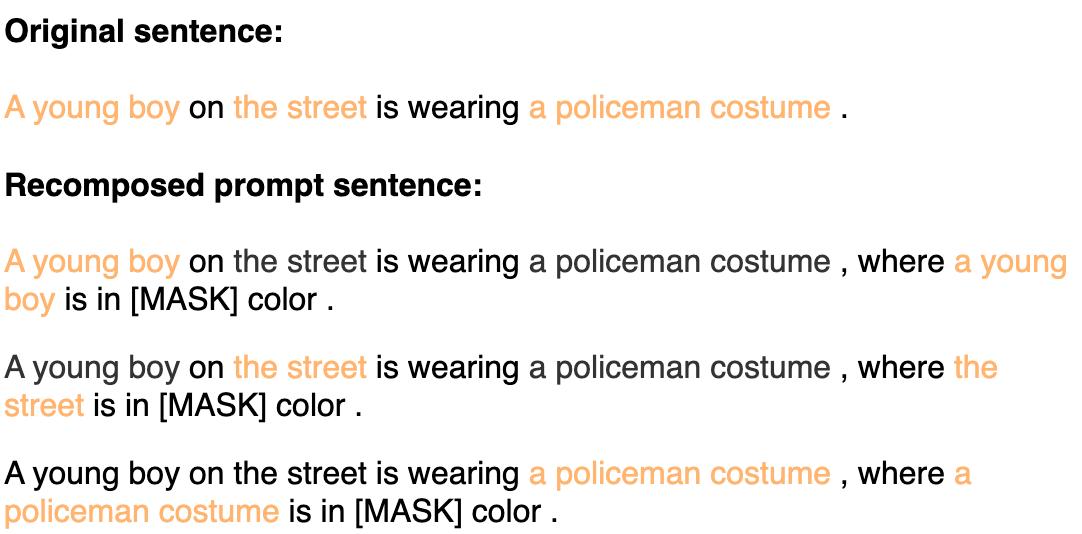}
    \caption{An example of CPT-adapted.}
    \label{fig:cpt-adapted}
\end{figure}

    We show an example of prompting CPT for phrase grounding in Figure \ref{fig:cpt-adapted}. The \textcolor[RGB]{255,181, 112}{yellow} words are the phrases we need to ground.

\end{document}